\pdfoutput=1

\documentclass[11pt]{article}

\usepackage[preprint]{acl}

\usepackage{times}
\usepackage{latexsym}

\usepackage[T1]{fontenc}

\usepackage[utf8]{inputenc}
\usepackage{graphicx}
\usepackage{microtype}

\usepackage{inconsolata}

\usepackage{titlesec}
\usepackage{hyperref}
\usepackage{multirow}
\usepackage{lipsum}
\usepackage{geometry}
\usepackage{multicol}
\usepackage{siunitx}

\usepackage{silence}
\newcommand{\psy}{\textsuperscript{*}}
\newcommand{\cs}{\textsuperscript{**}}

\title{A Multidimensional Framework for Evaluating Lexical Semantic Change with Social Science Applications}

\author{Naomi Baes\textsuperscript{\psy}~\;~Nick Haslam\textsuperscript{\psy}~\;~Ekaterina Vylomova\textsuperscript{\cs} \\ \textbf{\textsuperscript{\psy}} Melbourne School of Psychological Sciences \\ \textbf{\textsuperscript{\cs}} School of Computing and Information Systems \\ The University of Melbourne \\ \texttt{\{n.baes, nhaslam, vylomovae\}@unimelb.edu.au }}

\usepackage{booktabs} 
\usepackage{float}

\titleformat{\section}[block]{\filcenter\bfseries}{\thesection.}{1em}{}

\begin{document}
\maketitle
\begin{abstract}
Historical linguists have identified multiple forms of lexical semantic change. We present a three-dimensional framework for integrating these forms and a unified computational methodology for evaluating them concurrently. The dimensions represent increases or decreases in semantic 1) sentiment (valence of a target word’s collocates), 2) breadth (diversity of contexts in which the target word appears), and 3) intensity (emotional arousal of collocates or the frequency of intensifiers). These dimensions can be complemented by the evaluation of shifts in the frequency of the target words and the thematic content of its collocates. This framework enables lexical semantic change to be mapped economically and systematically and has applications in computational social science. We present an illustrative analysis of semantic shifts in \emph{mental health} and \emph{mental illness} in two corpora, demonstrating patterns of semantic change that illuminate contemporary concerns about pathologization, stigma, and concept creep.
\end{abstract}

\section{Introduction}

Lexical semantic change is defined by historical linguists as innovations that alter the meaning, but not the grammatical function, of a form \cite{Campbell:1999}. For instance, ``awesome'' once denoted the capacity to inspire awe, but its meaning has since been bleached to a general expression of approval. Computational linguists have made strides in developing distributional semantic methods \cite{boleda2020distributional} to detect semantic change \cite{Kutuzov:2018, Tahmasebi:2021, tang2018state} and its laws \cite{hamilton2016diachronic} as distinct from cultural shifts \cite{hamilton2016cultural}. 

Advances in deep learning since 2018 \cite{manning2022human} afford new ways to model semantic change processes. These innovations have facilitated the development of language models with sophisticated word embeddings or vector representations. As a result, word embeddings have evolved from count-based models \cite{jurafsky2024}, where words are represented by their co-occurrence frequency with other words, to prediction-based representations \cite{mikolov2013efficient,pennington2014glove}, where word vectors are iteratively learned as part of a language modelling task objective. The granularity of these representations shifted from \emph{type-level}, where each word has a single vector despite its usages, to \emph{token-based}, or contextualized representations \cite{Montanelli:2023, kutuzov2022contextualized}, where each word instance (token) has a vector, dynamically capturing shifts in meaning based on context. Lexical semantic relations can be detected by type- \cite{shwartz2016improving, vylomova-etal-2016-take} and token-level \cite{rogers2020primer} embeddings. 

Other work has started addressing the challenge of formalizing and understanding kinds of semantic change \cite{hengchen2021challenges}. Processes such as broadening \cite{Vylomova:2019,yuksel2021semantic}, metaphorization \cite{maudslay2022metaphorical}, and pejoration/amelioration \cite{fonteyn2021adjusting} have been modelled. Researchers have created methods to automatically disambiguate a word's pejorative usage from its non-pejorative use \cite{Dinu:2021}. Attempts have also been made to evaluate understudied classes of semantic change. Sentence representations from neural language models were used for hyperbole detection \cite{Schneidermann:2023}. Exaggerated language can be generated \cite{Tian:2021} and detected \cite{Kong:2020}, alongside metaphor \cite{badathala:2023}. Researchers have also evaluated semantic bleaching, whereby words lose elements of their meaning \cite{Luo:2019}, and found it to be triggered in contexts where an adverb premodifies a semantically similar adjective (e.g., “insanely jealous”). Nevertheless, there are a dearth of diachronic methods for evaluating lexical semantic change \cite{de2024survey}.

Despite advances in detecting and modelling lexical semantic change, there is a need for a unifying framework to integrate multiple dimensions of change. The present study addresses this gap by proposing a framework which synthesizes the theoretical insights of historical linguists about the many distinct forms of diachronic lexical semantic change \citep[e.g.,][]{bloomfield1933} and aligns them with the methodological sophistication of natural language processing. The comprehensive computational framework for evaluating lexical semantic change that emerges should be valuable for computational social scientists seeking to understand and model social and cultural change.

\section{Related Work}

\subsection{Forms of Lexical Semantic Change}

Historical linguists have developed several taxonomies of the forms of lexical semantic change \cite{Blank:1999, Bréal:1897, Ullmann:1962}, but \citeauthor{bloomfield1933}'s (\citeyear{bloomfield1933}) is one of the most well-established. Bloomfield described nine forms identified by earlier scholars: (1) narrowing: superordinate to subordinate, or when a meaning becomes more restricted (Old English \emph{mete} ‘all food’ > \emph{meat} ‘edible flesh’); (2) widening: subordinate to superordinate, or specific to general expansion of meaning (Middle English \emph{dogge} ‘dog of a specific breed’ > dog); (3) metaphor: the transfer of a name based on the associations of similarity or hidden comparison (Primitive Germanic  \emph{bitraz} ‘biting’, derivative of ‘I bite’ > \emph{bitter} ‘harsh of taste’), (4) metonymy: change based on the meanings’ proximity in space or time (Old English \emph{ceace} ‘jaw’ > \emph{cheek}); (5) synecdoche: the meanings are related as whole and part (pre-English \emph{stobo} ‘heated room’ > stove), (6) hyperbole: stronger to weaker meaning by overstatement (pre-French \emph{extonare} ‘to strike with thunder’ > to astonish; English borrowed \emph{astound}, \emph{astonish} from Old French); (7) meiosis:\footnote{Bloomfield (1933) refers to this class as litotes, but we use meiosis to reflect general understatement.} weaker to stronger meaning by understatement (pre-English \emph{kwalljan} ‘to torment’ > Old English \emph{cwellan} ‘to kill’); (8) degeneration: positive to negative connotation (Old English \emph{cnafa} ‘boy servant’ > \emph{knave}); (9) elevation: negative to positive connotation (Old English \emph{cniht} ‘boy, servant’ > \emph{knight}).

Bloomfield’s classes align closely with the forms of change identified in studies of denotational and connotational meaning \cite{Geeraerts:2010}. For denotational (referential) meaning, Geeraerts identifies (1) specialization, (2) generalization, (3) metonymy, and (4) metaphor. Specialization (semantic ‘restriction’ and ‘narrowing’) implies that the new meaning covers a subset of the old meaning's range; for generalization (or ‘expansion’, ‘extension’, ‘schematization’, ‘broadening’), the new range includes the old meaning.  Metonymy (here including synecdoche) is a “link between two readings of a lexical item based on a relationship of contiguity between the referents of the expression in each of those readings” \cite[p. 27]{Geeraerts:2010}. Conversely, metaphor is based on similarity. Geeraerts also identifies two forms of connotational meaning (i.e., the aspects of a word's meaning that are related to the writer or reader's emotions, sentiment, opinions, or evaluations): (1) pejorative and (2) ameliorative change (i.e., shift towards a more negative/positive emotive meaning). An example of pejoration is ‘silly’, which formerly meant ‘deserving sympathy, helpless’, but has come to mean ‘showing a lack of common sense’. Amelioration is shown by `knight' once meaning ‘boy, servant’.


\subsection{Expanding Concepts of Harm and Pathology}

Semantic change processes such as these may partly reflect cultural, social, and political shifts, and are of interest to social science researchers. One example is social psychological research on concept creep, the semantic expansion of harm-related concepts (e.g., abuse, bullying, mental illness, prejudice, trauma, violence; \citeauthor{Haslam:2016}, \citeyear{Haslam:2016}). Concept creep takes two forms: harm-related concepts have expanded ‘horizontally’ to cover a wider range of harms and ‘vertically’ to encompass less intense harms. It is theorized to be driven by rising cultural sensitivity to harm \cite{Furedi:2016, Wheeler:2019}, falling societal prevalence of harm \cite{levari2018, Pinker:2011}, and deliberate conceptual expansion by “opprobrium entrepreneurs” \cite{Sunstein:2018}. Concept creep is theorized to have mixed blessings \cite{Haslam:2020}, trivializing harms on one hand \cite{dakin:2023} and enhancing the recognition and redress of major harms on the other \cite{Tse:2021}.

Prior empirical work has evaluated concept creep in historical text corpora. Studies assessing horizontal expansion as increases in the broadening of harm concepts found that some concepts (e.g., addiction, bullying, trauma) have broadened within academic psychology \cite{Haslam:2021, Vylomova:2019, Vylomova:2021a}. Recent work evaluated the vertical form of concept creep, defined as the concept’s use in contexts of declining emotional intensity, and yielded mixed findings for anxiety, depression, grief, stress, and trauma \cite{Baes:2023, Baes:2023b, xiao:2023}.

Mental illness has become an increasingly salient term in society \cite{Haslam:2024}, partly due to the recent prioritization of mental health in global health policy \cite{WHO2021}. Critics have raised concerns that the rising prominence of mental health discourse is instigating problematic changes in how people conceptualize mental ill health. Some contend that concepts of mental illness have broadened so that everyday life is increasingly pathologized \cite{Brinkmann:2016, Horwitz:2007, Horwitz:2012}. Experiences that were once considered normal are now given diagnostic labels, such as using ‘depression’ to reference ordinary sadness \cite{Broer:2017}. Alternatively, it has been argued that terms like “mental health problems” are being normalized and broadened \cite{Sartorius:2007}, alongside increasing prevalence of mental illnesses. Some argue that concepts of mental illness are becoming less stigmatizing, although this question has only been addressed in surveys of public attitudes \citep[e.g., ][]{Schomerus:2022}, rather than in changes in word connotations. In view of the widespread speculation on the ways in which concepts of mental illness have changed historically and the lack of scientific evidence of these shifts, a systematic study of conceptual change in this domain is a priority.

\subsection{Our Original Contribution}

The present study aims to make three main contributions: (1) it proposes a multidimensional framework for evaluating lexical semantic change that economically integrates forms identified by historical linguists; (2) it develops a set of computational methodologies for evaluating change on these dimensions; and 3) it illustrates this computational framework by examining semantic shifts in concepts of mental health and mental illness to address cultural concerns about pathologization, normalization, and stigmatization. The study will therefore test if the framework can thoroughly illuminate how \emph{mental health} and \emph{mental illness} have changed their meanings in two corpora representing academic psychology and general US English text.\footnote{The source code is available here: \url{https://github.com/naomibaes/lexical_semantic_change_framework}}

\section{Method}
\subsection{Framework}
\label{sec:framework}
\begin{figure}[ht]
 \centering
 \includegraphics[width=\columnwidth]{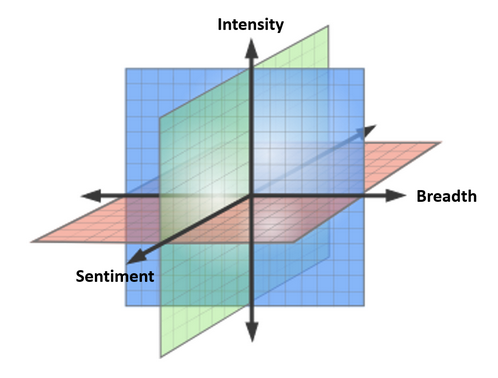}
    \caption{Three Major Dimensions of Semantic Change. }
    \label{fig:1}
\end{figure}

The proposed framework, illustrated in Figure~\ref{fig:1}, economically reduces classes of lexical semantic change identified by historical linguists \citep[excluding metaphor and metonymy;][]{Geeraerts:2010} to three dimensions. It recognizes that these classes represent opposed pairs of change types, each member corresponding to a pole on a single dimension. In essence, the framework reformulates six classes as three dimensions, allowing lexical semantic change to be quantified on three axes simultaneously rather than categorized into exclusive types. A recent survey paper \cite{de2024survey} has also classified semantic change as having three classes of characterizations related to a word's meaning becoming used in a more (1) pejorative or ameliorated sense (orientation), (2) metaphoric or metonymic context (relation), (3) abstract/general or more specific/narrow context (dimension). However, their theoretical framework does not consider hyperbole/litotes.

In our proposed framework, the \emph{Sentiment} dimension relates to whether the word acquires a more positive (‘elevation’, ‘amelioration’) or negative (‘degeneration’, ‘pejoration’) connotation. The \emph{Breadth} dimension relates to whether a word expands (‘widening’, ‘generalization’) or contracts (‘narrowing’, ‘specialization’) its semantic range. The \emph{Intensity} dimension relates to whether a word changes to refer to more emotionally or referentially intense phenomena (‘meiosis’) or less intense phenomena (‘hyperbole’).  Table {\ref{tab:1}} summarizes how the three dimensions map onto the classes of lexical semantic change as well as the two proposed forms of concept creep. 

\begin{table*}[!ht]
\centering
\begin{tabular}{lp{65mm}p{65mm}}
\toprule 
\textbf{Dimension} & \textbf{Rising}  & \textbf{Falling}\\
\midrule 
Sentiment & Elevation \cite{bloomfield1933}; Amelioration \cite{Ullmann:1962} & Degeneration \cite{bloomfield1933}; Pejoration \cite{Ullmann:1962}\\
Breadth & Widening \cite{bloomfield1933, Ullmann:1962};  Generalization of meaning \cite{Blank:1999}; Horizontal Creep \cite{Haslam:2016}* & Narrowing \cite{bloomfield1933, Ullmann:1962}; Specialization of meaning \cite{Blank:1999} \\
Intensity & Meiosis \cite{bloomfield1933} & Hyperbole \cite{bloomfield1933}; Vertical Creep \cite{Haslam:2016}*\\
\bottomrule 
\end{tabular}
\caption{Dimensions of Lexical Semantic Change and their associated forms. * = specific to harm-related concepts.}
\label{tab:1}
\end{table*}

The three proposed dimensions align with established dimensions in other domains. For example, Sentiment and Intensity resemble the two primary dimensions of human emotion, Valence and Arousal \cite{Russell:2003}, and two primary dimensions of connotational meaning, Evaluation (e.g., “good/bad”) and Potency (e.g., “strong/weak”) \cite{Osgood:1975}, both of which have been shown to have cross-cultural validity. Although our dimensions capture the primary forms of lexical change, we argue that they can be complemented by evaluation of changes in a word’s salience (i.e., relative frequency of use) and its \emph{thematic content} (i.e., shifts in the specific contexts in which the word is used). These dimensions may reflect psychological, sociocultural, or cultural forces that contribute to or result from semantic change \cite{Blank:1999}. Our case study of \emph{mental health} and \emph{mental illness} illustrates how attention to salience and thematic content enrich the characterization of semantic change that the three primary dimensions provide. We now turn to the details of that case study, including the computational methodologies for evaluating these dimensions. Future implementations of our three-dimensional framework are likely to include technical refinements of these methodologies. Those employed in the case study simply demonstrate one way to implement it using interpretable techniques.

\subsection{Sentiment}
\label{sec:sentiment}

The sentiment of the target concepts (\emph{mental health} and \emph{mental illness} and the control concept \emph{perception}) was evaluated using valence norms from \citeauthor{Warriner:2013} (\citeyear{Warriner:2013}), which provide valence ratings for 13,915 English lemmas collected from 1,827 United States residents, ranging from low valence (1: feeling extremely “unhappy”, "despaired") to high valence (9: feeling extremely “happy”, “hopeful"). See Appendix~\ref{sec:appendix_A} for more information regarding the valence ratings. Collocates of each target concept were extracted within a ± 5-word context window \cite{Agirre:2009} and matched to the Warriner et al. norms which showed adequate coverage for the psychology corpus but poorer coverage for the general corpus (“mental\_health”: psychology = 84\%; general = 50\%; “mental\_illness”: psychology = 83\%;general = 48\%; “perception”: psychology = 84\%; general = 39\%). Annual counts of Warriner-matched collocates for each target concept were then extracted from the lemmatized corpora, which showed few occurrences due to few appearances of texts containing targets  before 1990 in the general corpus (see Appendix~\ref{sec:appendix_B}). Therefore, analyses excluded general texts before 1990. The annual sentiment score for each concept was computed by weighting the valence rating for each collocate by its annual appearances, standardized by the total number of (matched) collocates in the respective year. The index represents the mean valence of terms [1,9] collocating with target concepts, where higher scores indicate higher valence.
    

\subsection{Breadth}
\label{sec:breadth}
The semantic broadening of the target concept was evaluated as the average inverse cosine similarity between the sentence level embeddings containing the target term. Our method adapts previous work \cite{Vylomova:2019, Vylomova:2021a} by replacing type-level word embeddings with contextualized sentence-level embeddings. Given that this breadth measure resembles the Semantic Textual Similarity (STS) task \cite[the degree to which two sentences are semantically equivalent to each other]{cer:2017}, to select the optimal model we compared the sentence similarity scores, from corpus samples, of models that have shown good performance for encoding sentences. Many of the original Sentence-BERT models \cite{Reimers&Gurevych:2019} with good scores on semantic textual similarity benchmarks \cite{tsukagoshi2022comparison, Reimers&Gurevych:2019} are deprecated, therefore we examined and compared three public pre-trained models that currently excel in encoding sentences,\footnote{\url{https://www.sbert.net/docs/pretrained_models.html}} from the sentence transformers library. See Appendix \ref{sec:appendix_C} for more information regarding model selection (\ref{sec:model_select}), comparison (\ref{sec:model_compar}) and results (\ref{sec:model_results}). The pre-trained model used in the present study\footnote{"all-mpnet-base-v2" from Hugging Face, sentence-transformers: \url{https://huggingface.co/sentence-transformers/all-mpnet-base-v2}} performed best on detecting \emph{semantic} information and encoding sentences for 14 diverse tasks from different domains.

To compute the breadth score, relevant texts were extracted from our corpora. Inspecting their frequencies showed that it was acceptable to sample 50 texts from each five-year interval.\footnote{Appendix \ref{sec:breadth_measure} explains interval selection.} Thus, we randomly and uniformly sampled up to 50 sentences per interval and repeated the procedure 10 times to reduce sampling noise. These sentences were then passed to the sentence transformer model, "all-mpnet-base-v2" (where MPNET means Masked Permuted Language Modeling Network), to be tokenized and to encode embeddings representing their semantic characteristics. Cosine distance was computed for each pair of sentence vectors by inverting the similarity scores (1 - cosine similarity). The final breadth metric [0,1] was calculated by averaging scores across samples in each interval. Higher scores indicate greater breadth (\emph{dis}similarity) between sentence vectors.
%

\subsection{Intensity}
\label{sec:intensity}
Changes in the intensity of the concepts were evaluated in two ways. First, we computed an arousal index, adapting a previously established procedure \cite{Baes:2023, Baes:2023b, xiao:2023}. In an equivalent manner to the sentiment analysis, we examined the collocates of each concept and computed a weighted average annual ratings, using Warriner et al.’s arousal norms that range from low arousal (1: feeling "calm", "unaroused" while reading the lemma) to high arousal (9: feeling "agitated", "aroused"). See Appendix~\ref{sec:appendix_A} for more information regarding arousal ratings. The annual arousal score for each concept was calculated by weighting the arousal rating for each collocate by its total number of appearances in each year and normalizing it by the total (matched) collocate count for the respective year. The index represents the mean arousal of terms [1,9] collocating with target concepts, where higher scores indicate higher arousal. 

Second, we developed a new index to directly capture shifts in a concept’s intensity. Instead of examining the arousal of its collocates (regardless of their order), it examined the occurrence of intensifying expressions that directly modify it. If a concept increasingly appears with an intensifying modifier, it can be inferred that its unmodified meaning has become less intense. We developed a new “intensifier index” which evaluates the relative frequency with which 11 adjectival modifiers (“great”, “intense”, “severe”, “harsh”, “major”, “extreme”, “powerful”, “serious”, “devastating”, “destructive”, “debilitating”) preceded “mental health” and “mental illness”. De-adjectival adverbs from \citeauthor{Luo:2019} (\citeyear{Luo:2019}) were considered but most were not sufficiently general (e.g., “devastating”, “excruciating”, “vicarious”). We used the dependency-parsed corpora (see Section \ref{sec:preprocessing}) to compute the proportion of instances of each target concept that has any of the 11 terms as its adjective modifier.

\subsection{Thematic content}
\label{sec:theme}
Thematic content was evaluated using a top-down approach. The theme of interest was pathology given concerns raised by critics about the pathologization of \emph{mental health} and \emph{mental illness} \cite{Brinkmann:2016, Horwitz:2007, Horwitz:2012}. We used a pathologization dictionary developed by \citeauthor{Baes:2023} (\citeyear{Baes:2023}) to compute the pathologization index. This approach can be used to construct dictionaries for other themes of interest. First, we generated unambiguously disease-related words with restricted range in meaning: “clinical”, “disorder”, “symptom”, “illness”, “pathology”, and “disease”. Next, their forward word associations (participant responses to each disease-related word) drawn from the English Small World of Words project \cite{DeDeyne:2019} were listed and duplicates were removed. We filtered the list for terms reflecting pathologization (i.e., to view or characterize as medically or psychologically abnormal), leaving 17 terms: “ailment”, “clinical”, “clinic”, “cure”, “diagnosis”, “disease”, “disorder”, “ill”, “illness”, “medical”, “medicine”, “pathology”, “prognosis”, “sick”, “sickness”, “symptom”, “treatment”. Following \citeauthor{Baes:2023} (\citeyear{Baes:2023}), we computed the pathologization index by dividing appearances of the 17 terms in the target concept’s collocates (±5-word context window) in a specific year by the total number of collocates in that year. 

\subsection{Salience}
\label{sec:salience}
Salience was computed as the concept’s annual relative frequency, using the raw corpora versions.

\section{Materials}
\subsection{Corpora}

Two corpora were chosen for their historical length, their magnitude, and their texts. The psychology corpus contained 143,575,773 tokens from 871,344 abstracts from 875 (Scimago indexed) psychology journals, ranging from 1930 to 2019, sourced from E-Research and PubMed databases \cite{Vylomova:2019}. The journal set was distributed across all subdisciplines of psychology. The final corpus of psychology abstracts was limited to 1970-2016 due to the relatively small number of abstracts outside this period \cite{Vylomova:2019}, yielding 129,980,596 tokens from 793,942 abstracts.

The second corpus is a combination of two related corpora: the Corpus of Historical American English \cite[1810-2009]{Davies:2010} and the Corpus of Contemporary American English \cite[1990-2019]{Davies:2008}. Academic texts were excluded to avoid any potential overlap with psychology articles. After merging the two corpora, containing 115,000 everyday publications and >500,000 contemporary texts, the combined corpus was processed following recommendations from \citeauthor{alatrash:2020} (\citeyear{alatrash:2020}) to maintain data integrity.\footnote{See Appendix \ref{sec:appendix_D} for a comprehensive explanation.} The current study restricted the corpus period from 1970 to 2016, using 501,415,577 tokens from 244,552 texts (books: 23,855 fiction, 1,498 non-fiction; 88,641 magazines; 73,557 newspapers; 40,036 spoken language; 16,965 TV shows). 

\subsection{Preprocessing}
\label{sec:preprocessing}

Analyses required three versions of the corpora: (1) a raw cleaned version transforming target concepts to single noun tokens (Section~\ref{sec:salience} and \ref{sec:breadth} and ~\ref{sec:intensity}); (2) a lemmatized version (Section~\ref{sec:sentiment}, \ref{sec:intensity}, and \ref{sec:theme}); and (3) a dependency parsed version (Section~\ref{sec:intensity}).
The first version, including punctuation, uppercasing, and numbers, was used for all analyses after transforming multiword target concepts into single tokens (e.g., “mental health” > “mental\_health”) using case sensitive matching. The lemmatization pipeline included tokenization, part-of-speech tagging (skipping tokens with uninformative tags: punctuation, symbols, spaces, numbers), removing stop words (uninformative words like “the”), and lemmatization using spaCy.\footnote{\url{https://spacy.io/}} For dependency parsing we used the raw corpora to provide more contextual information for the model to better understand relationships between words. The English Transformer model\footnote{``en\_core\_web\_trf'' (roberta-base) from Spacy was used as it demonstrates the highest accuracy on 13 evaluation tasks: \url{https://spacy.io/models/en\#en_core_web_trf.}} was used to preprocess the corpus with a high performance computing system \cite{lafayette2016spartan}.

\subsection{Target Concepts}
Two terms were chosen to analyze levels of semantic change \cite{hamilton2016cultural}: \emph{mental\_health} and \emph{mental\_illness}. We also ran control analyses using the neutral term, \emph{perception}, for which a fixed rate of change was expected and which demonstrated a steady rise in relative frequency starting around 1945 in the Google Ngram Viewer.\footnote{\url{https://books.google.com/ngrams/info}}

\subsection{Statistical Analysis}
Linear regression analyses were performed to test the statistical significance of historical trends in the semantic indices \cite{Jebb:2015}. Ordinary least squares served as the primary estimator, the secondary one being a generalized least squares estimator to account for auto-correlated residuals (Durbin-Watson test: \emph{p} < .05). Coefficients, standard errors and confidence intervals were standardized using the betaSandwich package \cite{pesigan2023betadelta}, employing Dudgeon's (\citeyear{dudgeon2017some}) heteroskedasticity-consistent estimator approach (HC3), ideal for extracting estimates for nonnormal data and small sample sizes \cite{dudgeon2017some}. The code is publicly available.{\footnote{\url{https://osf.io/4d7ur/}}



\section{Results}

\textbf{Sentiment}: The linear regression models mostly show decreasing trends for the valence index. Figure~\ref{fig:2} shows a significant declining trend in the valence of words used in the context of \emph{mental health} in the psychology corpus and the general corpus. For \emph{mental illness}, the valence index shows a decreasing trend in psychology, and an increase in the general corpus. The valence of \emph{perception} only shows a decreasing trend in the general corpus.

\begin{figure}[ht]
 \centering
 \includegraphics[width=\columnwidth]{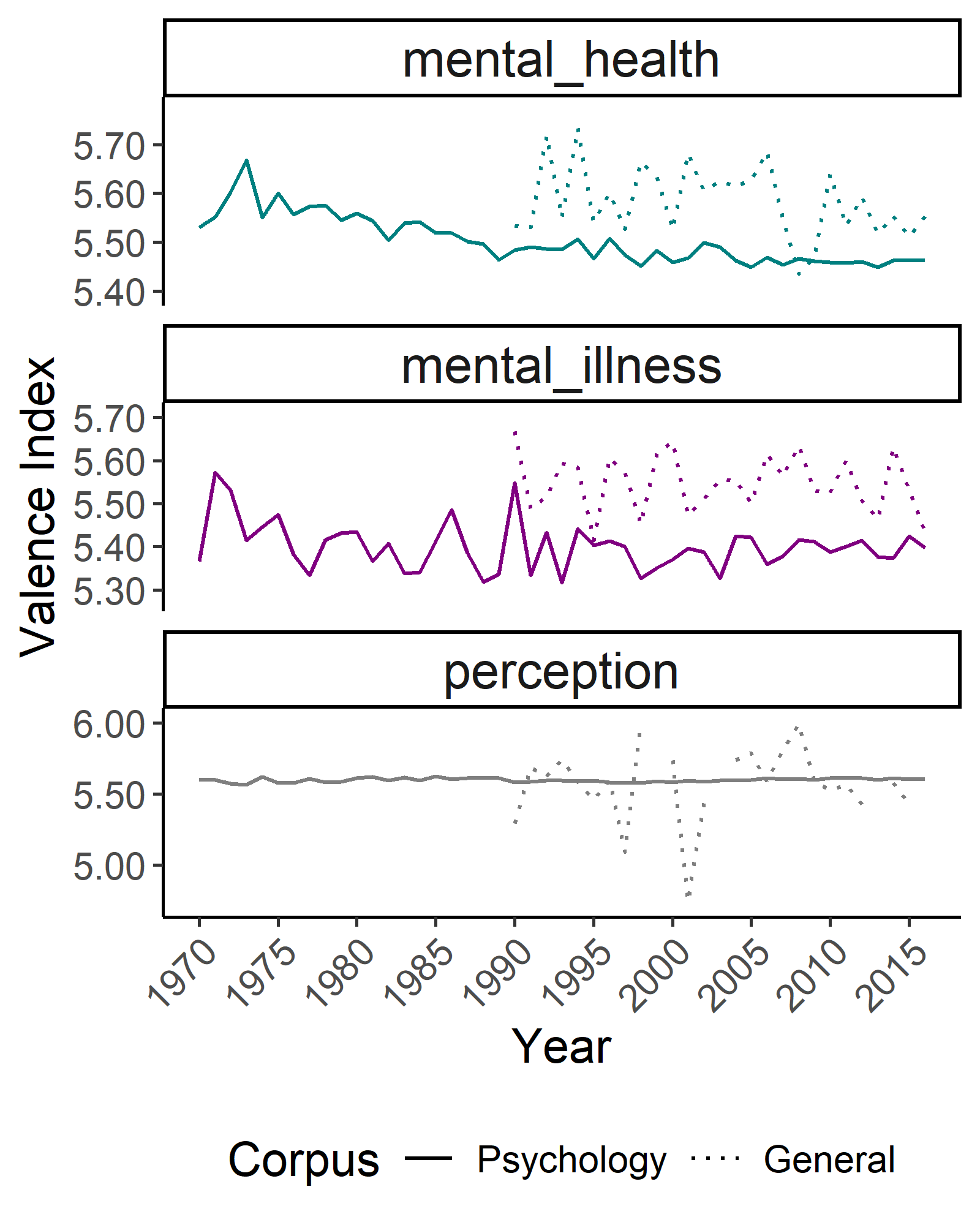}
    \caption{Valence index over the study period (1970-2016).}
    \label{fig:2}
\end{figure}

\textbf{Breadth}: The linear regression models testing the trend for the cosine distance of sentential contexts containing targets show significant increasing trends for \emph{mental health}, \emph{mental illness} and \emph{perception} in the psychology corpus, reflecting greater sentence diversity, with a decrease for \emph{mental health} and an increase for \emph{perception} in the general corpus, as shown in Figure~\ref{fig:3}.

\begin{figure}[ht]
 \centering
 \includegraphics[width=\columnwidth]{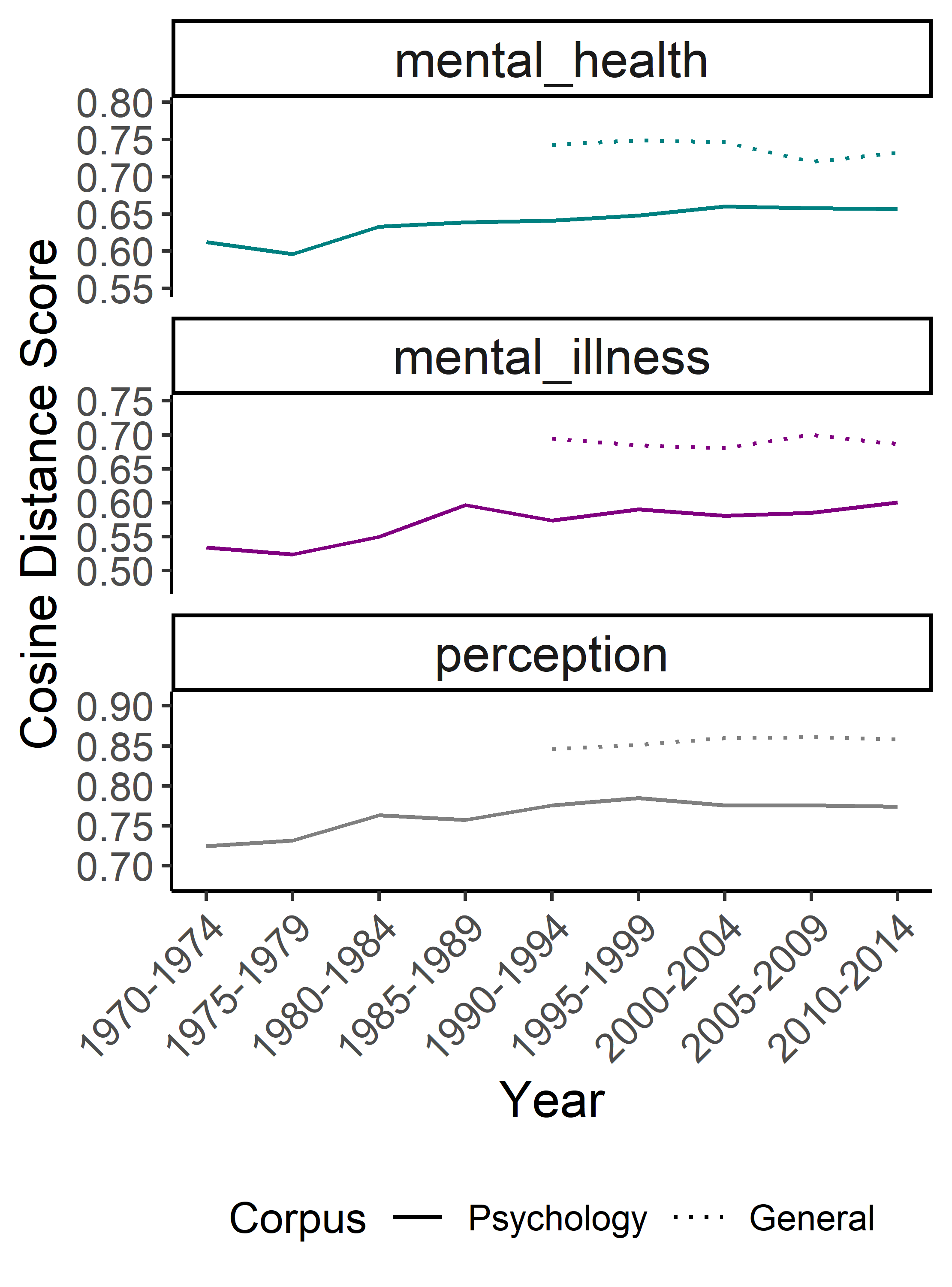}
    \caption{Breadth score over five-year intervals (1970-2014). }
    \label{fig:3}
\end{figure}

\textbf{Intensity}: Figure~\ref{fig:4} shows the significant rise and fall in the use of intensifiers to modify \emph{mental illness} in the psychology corpus, but no trend in the general corpus. Examining the top ranked adjective modifiers in each decade (Table~\ref{tab:table_E2} and Table~\ref{tab:table_E5} in Appendix~\ref{sec:appendix_E}) reveals that “severe”, “serious”, “major”, “chronic” come to be more associated with \emph{mental illness} from the 1990s onwards. Although \emph{mental health} is not frequently modified by intensifiers, as expected, “poor” and “positive” remain closely associated with it across the decades, with “maternal” becoming more associated with \emph{mental health} from the 1990s onwards. Despite demonstrating a significant increase in its intensifier index in the psychology corpus, \emph{perception} does not display intensifiers among its top adjective modifiers.

\begin{figure}[ht]
 \centering
 \includegraphics[width=\columnwidth]{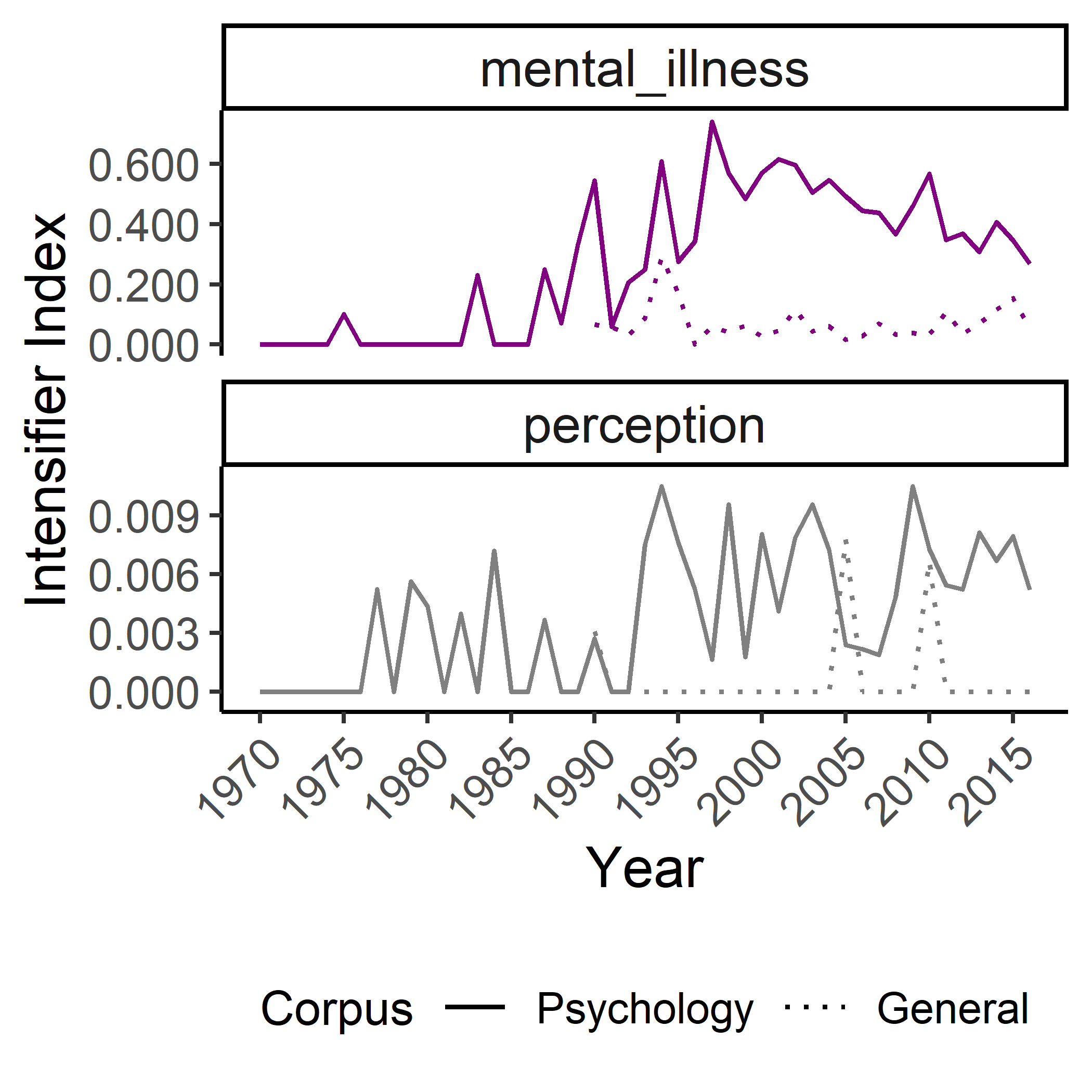}
    \caption{Intensifier index for \emph{mental illness} over the study period (1970-2016). }
    \label{fig:4}
\end{figure}

Figure~\ref{fig:5} shows a significant increasing trend in the intensity (arousal index) of \emph{mental health}-related words in both corpora. For \emph{mental illness} and \emph{perception}, the index increases significantly for the psychology corpus and only shows an increasing trend for \emph{perception} in the general corpus.

\begin{figure}[ht]
 \centering
 \includegraphics[width=\columnwidth]{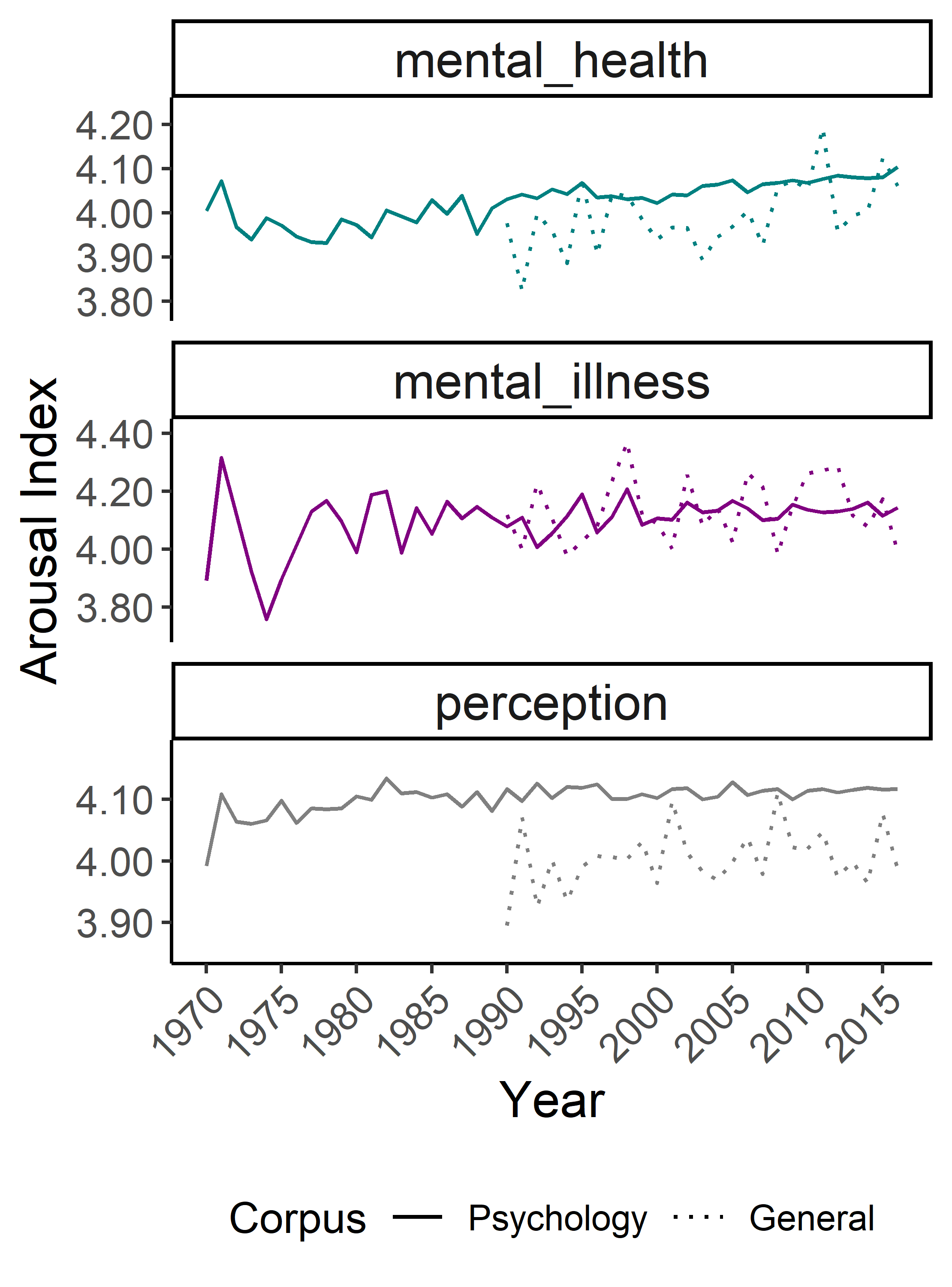}
    \caption{Arousal index over the study period (1970-2016). }
    \label{fig:5}
\end{figure}

\textbf{Thematic content}: The target concepts, \emph{mental health} and \emph{mental illness}, and the control \emph{perception}, become significantly more associated with pathology-related terms in the psychology corpus, and for all targets except for \emph{mental health} in the general corpus, as shown in Figure~\ref{fig:6}. Inspecting the top ten ranked collocates for the main target terms (see Appendix \ref{sec:appendix_F}) shows the presence of only two of the 17 pathology-related terms in psychology and the general corpus (“disorder” and “treatment”), and no pathology-related terms among the top ranked collocates for the control. The diversity of terms among the top ranked collocates for \emph{mental health} and \emph{mental illness} indicate that more themes are present in the semantic space. 

\begin{figure}[ht]
 \centering
 \includegraphics[width=\columnwidth]{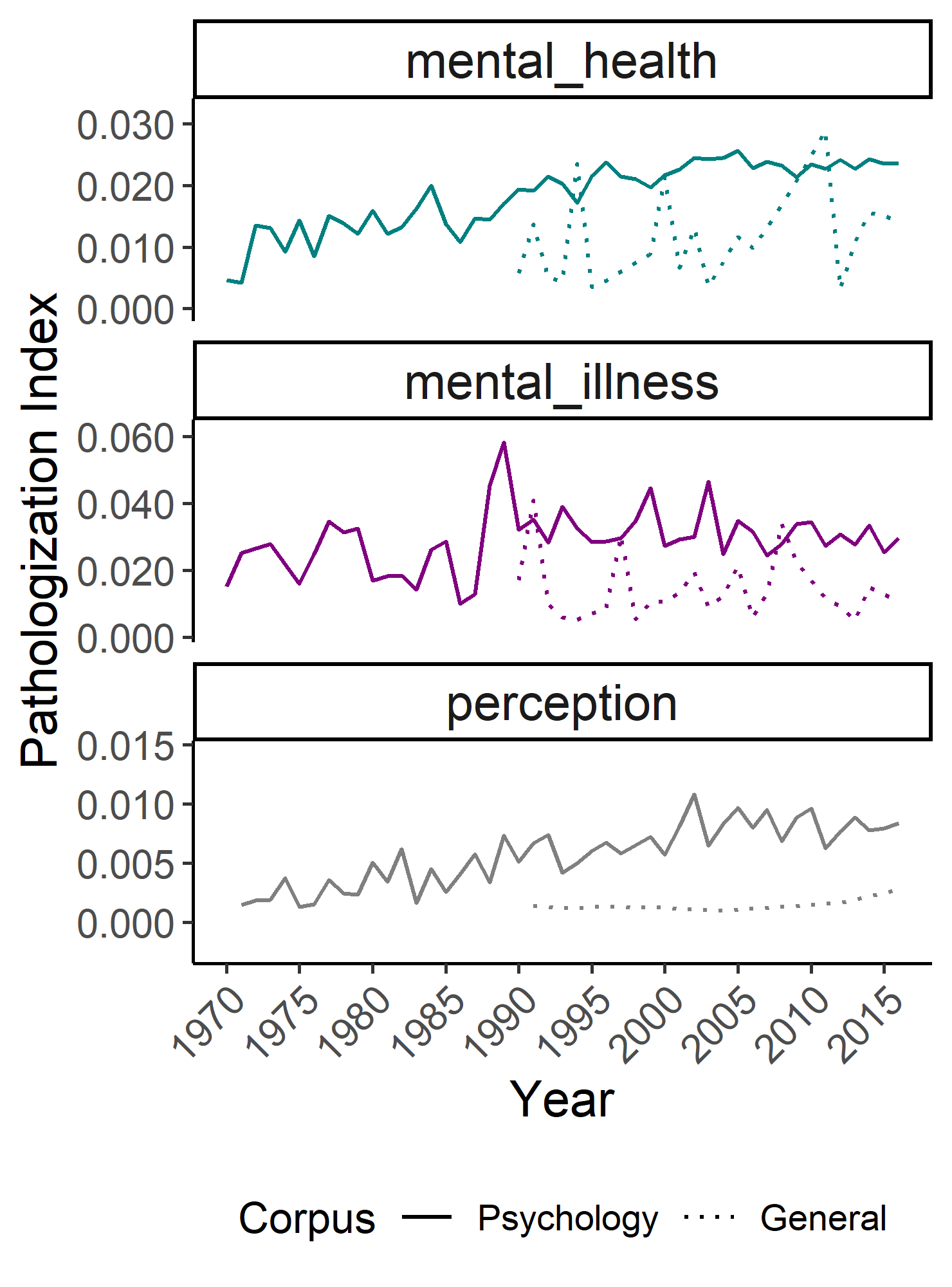}
    \caption{Pathologization index over the study period (1970-2016). }
    \label{fig:6}
\end{figure}

\textbf{Salience}: Figure~\ref{fig:7} illustrates that the relative frequencies rise significantly for both target concepts, \emph{mental health} and \emph{mental illness}, in both corpora. The relative frequency of \emph{perception} increases significantly in the psychology corpus and shows relatively stability in the general corpus.

\begin{figure}[ht]
 \centering
 \includegraphics[width=\columnwidth]{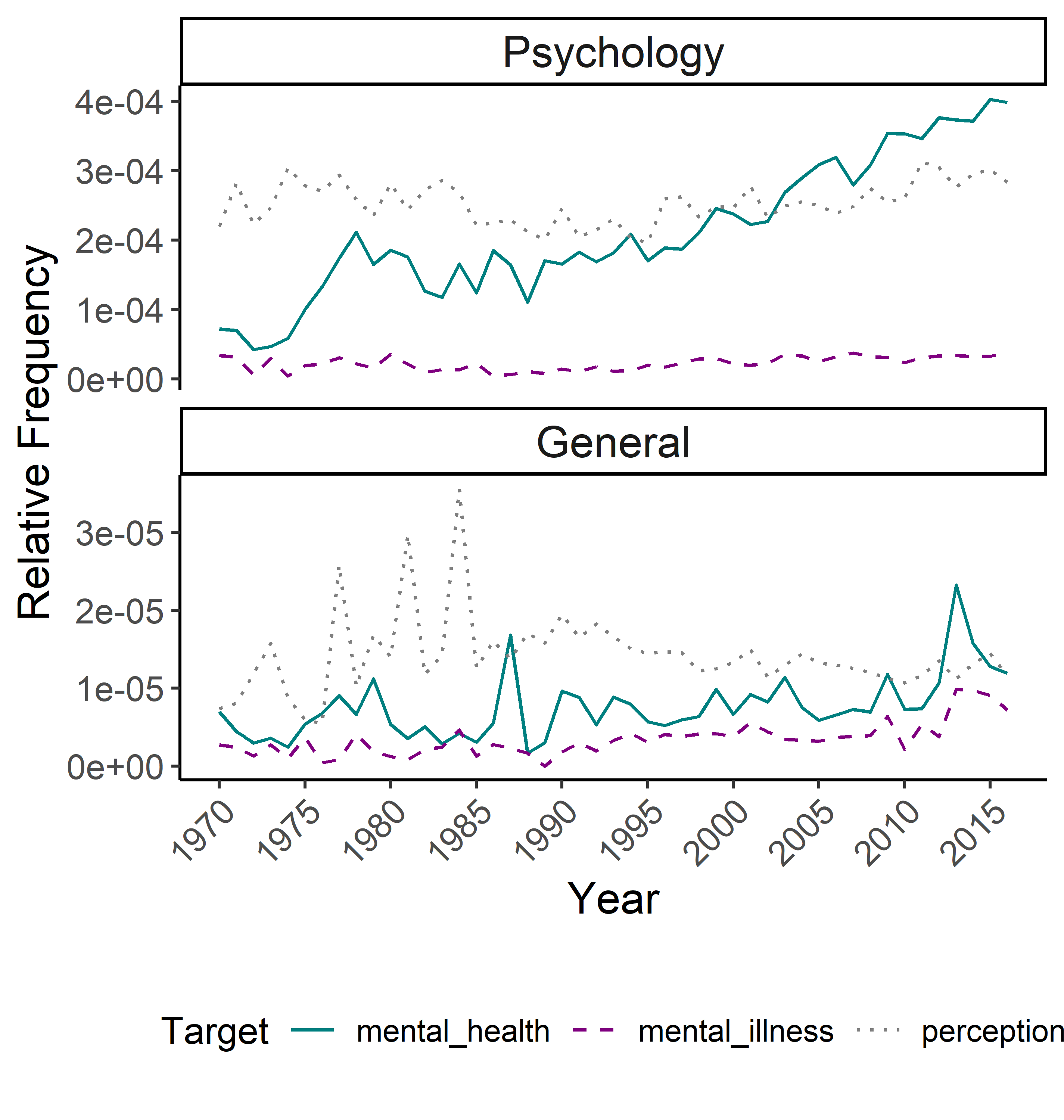}
    \caption{Normalized term frequencies for the general and psychology corpora (1970-2016).}
    \label{fig:7}
\end{figure}

The significance of the trends was determined by examining standardized beta coefficients and their associated standard errors (see Table~\ref{tab:table_AG3}). As shown in Appendix \ref{sec:appendix_G}, the strongest effect sizes can be observed for the two target terms with breadth (both corpora), valence (decreasing for psychology and increasing for the general corpus), and for \emph{mental illness} with intensity (both corpora). According to the Adjusted $R^2$ values in Tables~\ref{tab:table_AG1} and~\ref{tab:table_AG2}, with a few exceptions year has more explanatory power predicting the semantic indices for the target concepts than for the control concept.

\section{Discussion}

The present study implemented, for the first time, a new framework for evaluating lexical semantic change. Rather than assessing a single dimension of change or classifying it into a specific taxonomic category, the framework enables the concurrent evaluation of multiple dimensions of semantic change, each corresponding to a well-established dimension of referential or affective meaning. Evaluating semantic change along these dimensions simultaneously allows complex patterns of change to be disentangled and characterized, with possible applications in social science research.

The case study demonstrated a suite of computational methodologies for evaluating the framework’s dimensions of change in an examination of \emph{mental health} and \emph{mental illness} motivated by social scientific research questions. Theorists working in sociology, psychology, psychiatry, and related fields have speculated on recent cultural shifts in these concepts, relying on overlapping and sometimes ill-defined notions of medicalization \cite{Broer:2017, Hofmann:2016}, pathologization \cite{Brinkmann:2016, Frances:2013}, psychiatrization \cite{Beeker:2021, Paris:2020}, and stigmatization \cite{Sartorius:2007, Schomerus:2022}. Little research has investigated these proposed trends or attempted to characterize them systematically. Our case study documents how a rigorous characterization of these conceptual changes might be conducted. Its findings point to the complexity of these changes, which would remain hidden had they been evaluated on a single dimension.

Regarding sentiment, we found paradoxical trends. Sentiment toward \emph{mental illness} became more positive in the general corpus, supporting suggestions of destigmatization in the culture at large (e.g., \citeauthor{Schomerus:2022} \citeyear{Schomerus:2022}), while sentiment toward \emph{mental health} and \emph{mental illness} became more negative in the psychology corpus and for \emph{mental health} in the general corpus. In the general corpus, \emph{mental health} came to be used in narrower contexts. Nevertheless, the consistent rising trends for semantic breadth in the psychology corpus support previous claims of expanding meanings or horizontal concept creep \cite{Brinkmann:2016, Horwitz:2007, Horwitz:2012} in academic psychology.

Furthermore, the analysis of intensity yielded clear patterns of change. The target concepts rose on the arousal index in psychology, indicating that although only the semantic contexts for \emph{mental health} increased in valence, both \emph{mental health} and \emph{mental illness} have become more emotionally animated or agitated. Only \emph{mental health} showed no arousal trend in the general corpus. There was also evidence that \emph{mental illness} has increasingly become more and then less modified by intensifier adjectives, in the psychology corpus, possibly in response to vertical concept creep, where the concept’s meaning is stretched to refer to less severe phenomena \cite{Haslam:2016} which may lead people to intensify the target concept (e.g., where \emph{mental illness} comes to be modified as “serious” or “severe”) to distinguish it from more expansive usages. This increase in severity modifiers and arousal may both reflect the same rising concern with and problematization of mental illness and health. 

Finally, the tendency for the target concepts to become more associated with pathology-related terms (apart from for \emph{mental health} in the general corpus) supports claims of rising pathologization \cite{Brinkmann:2016}. Notably, \emph{mental illness} was most pathologized. Furthermore, the increase in relative frequency of the target concepts in both corpora is evidence of their rising cultural salience in psychology and the general domain. All indices for the control target, \emph{perception}, showed significant trends in at least one corpus.

In sum, the multi-dimensional analysis suggests that in recent decades, as discourse on mental health and illness has become more prominent (supported by our salience index), concepts of mental health and illness have not so much de-stigmatized (sentiment) but have instead inflated (breadth) and become a growing focus of social concern and problematization (intensity) and increasingly seen through a medical lens (pathologization).

\section{Conclusion}

The current study presented a new computational framework that can be applied in the social sciences. Our contributions lie in (1) proposing a multidimensional framework to evaluate lexical semantic change in a way that economically integrates forms identified by historical linguists; (2), developing a set of computational methodologies to evaluate change on the newly proposed semantic dimensions; and (3) illustrating the computational framework by examining how \emph{mental health} and \emph{mental illness} have changed their meanings in two corpora, implying that the concepts are increasingly inflated, problematized and pathologized. The investigation illuminates the complexity of semantic and cultural change and provides new tools for studying them.

\newpage
\section{Limitations}

Limitations inspire future directions. The procedures employed in the present study are simply a first implementation of the framework. Future research should refine its computational methodology by enhancing or replacing procedures with more robust or sensitive alterations. While the Warriner norms data we used (i) follows a rigorous and reliable rating procedure, (ii) are highly interpretable and (iii) have high face validity, future work might consider alternative methods in addition to closed-vocabulary approaches \cite{eichstaedt2021closed}. The current method could be compared against publicly-available BERT-based models fine-tuned for sentiment analysis \cite{goworek:2024}, the VADER (a rule-based sentiment analysis tool; \citeauthor{hutto:2014}, \citeyear{hutto:2014}), or other sentiment-emotion lexica \citep{boyd-graber-etal-2022-human, mohammad:2018}. Ideally, the approach will capture the nuanced sentiment contributions of the target word, which averaging the sentiment of contexts fails to capture \cite{goworek:2024}. Robustness checks should be conducted on new methods by comparing its convergent validity against the existing one to evaluate the extent to which the alternative method correlates when applied to the same dataset. In addition, because the target term's semantic broadening is operationalized as the cosine \emph{dis}similarity of the target's sentential contextual usages, it only differentiates between quantitatively (not qualitatively) different meanings. Future work should introduce more fine-grained follow-up analyses by, for example, identifying hypernymy or using state-of-the-art word in context (WiC) models, like XL-LEXEME \cite{cassotti:2023}, which beats GPT-4 on the WiC task and BERT, mBERT, XLM-R on the graded change detection task \cite{periti2024systematic}. It should also introduce a diachronic analysis to examine if the target's prototypical meaning has been diluted/intensified.

Additionally, while the present study includes a neutral control term, future work should evaluate how to (semi)automatically identify baseline semantic change in the global corpus (a stability axis), to normalize the semantic change of the target concepts against. A control condition where no change of meaning is expected could also be set up \cite{Dubossarsky:2017} using a chronologically shuffled corpus so that the assumed changes become uniform and any change is an artefact (reflects random "noise", not variation in time). To better capture themes, future work should develop a bottom-up, not a top-down dictionary-based, approach by using topic modeling or clustering contextualized word embeddings \cite{montariol:2021} and evaluating the target’s proximity to the centroid of the semantic category cluster. These methods might reveal senses or domains without imposing a dictionary on the semantic space. It will also be crucial to consider LLM approaches for lexical semantic change \cite{wang:2023}.

With regard to substantive studies, it will be important to make a general case for the framework by, ideally, finding an existing data set that includes annotated examples of semantic change for evaluation and estimation of the recall/coverage of the methods. In addition, our findings should be extended by applying the framework to a wider assortment of mental health-related concepts such as diagnostic terms (e.g., anxiety, depression, autism, obsessive-compulsive disorder, schizophrenia, attention-deficit hyperactivity disorder). Characterizing how specific diagnoses have altered their meanings in a differentiated, multi-dimensional manner will illuminate historical changes that have only been the focus of theoretical speculation and qualitative research to date 
\citep[e.g.,][]{Brinkmann:2016, Horwitz:2007, Horwitz:2012, Parrott:2023}. Future research can also capitalize on the new framework to explore possible causal relationships between dimensions, such as whether rising salience drives conceptual broadening \cite{Haslam:2021}, whether rising breadth of mental illness-related concepts drives improvements in sentiment (a destigmatization process), and whether trade-offs exist (e.g., rising breadth may lead to shifts in intensity). Studies already point to related laws of semantic change, finding that sentiment change is associated with semantic change \cite{goworek:2024}. Future studies should conduct fine-grained analyses on semantic shifts in discourse around mental health to examine how online group dynamics and macro social and cultural shifts \citep[e.g., prevailing stereotypes and stigma towards social groups; see][]{garg2018word, charlesworth2024mechanisms, durrheim2023using} contribute to observed semantic shifts and possibly the social transmission of mental disorders, shown in adolescent peer networks; \citet{alho2024transmission}. Ideally studies will be conducted with many corpora (e.g., news, social media) with high frequencies of the target terms.

\section{Ethics Statement}

We do not identify any foreseeable risks or potential for harmful use of our work. Analyses use licensed data that are openly accessible for academic purposes, ensuring transparency and accountability.

\section*{Acknowledgements}

We thank the three anonymous ACL reviewers for their valuable feedback which substantially improved the paper. Our gratitude also extends to Professor Charles Kemp and Professor Yoshihisa Kashima for their guidance and feedback on earlier versions of the framework at PhD committee meetings and to Lea Frermann, Filip Miletić and Andrey Kutuzov for indirect recommendations which benefited the work, and to Zheng Wei Lim for helping me troubleshoot this airXiv submission. This research is supported by Australian Research Council Discovery Project DP210103984 and by an Australian Government Research Training Program Scholarship.


\bibliography{anthology,custom}

\newpage

\appendix
\section{Appendix A}
\label{sec:appendix_A}

To elaborate on what a word being low or high "arousal" or "valence" means, \citet{Warriner:2013} defined them in the following way when (valid) participants made direct judgements of the large sample of words on the measured attributes (\emph{n} = 419: valence; \emph{n} = 448: arousal; 16-87 years; majority were female (60\%), English native language speakers, held a college degree):

\begin{itemize}
    \item {\bf{Valence}}: \emph{"You are invited to take part in the study that  [...] concerns how people respond to different types of words. You will use a scale to rate how you felt while reading each word. [...] The scale ranges from 1 (happy) to 9 (unhappy). At one extreme of this scale, you are happy, pleased, satisfied, contented, hopeful. When you feel completely happy you should indicate this by choosing rating 1. The other end of the scale is when you feel completely unhappy, annoyed, unsatisfied, melancholic, despaired, or bored. You can indicate feeling completely unhappy by selecting 9. The numbers also allow you to describe intermediate feelings of pleasure, by selecting any of the other feelings. If you feel completely neutral, neither happy nor sad, select the middle of the scale (rating 5)."}

    \item {\bf{Arousal}}: \emph{“You are invited to take part in the study that [...] concerns how people respond to different types of words. You will use a scale to rate how you felt while reading each word. [...] The scale ranges from 1 (excited) to 9 (calm). At one extreme of this scale, you are stimulated, excited, frenzied, jittery, wide-awake, or aroused. When you feel completely aroused you should indicate this by choosing rating 1. The other end of the scale is when you feel completely relaxed, calm, sluggish, dull, sleepy, or unaroused. You can indicate feeling completely calm by selecting 9. The numbers also allow you to describe intermediate feelings of calmness/arousal, by selecting any of the other feelings. If you feel completely neutral, not excited nor at all calm, select the middle of the scale (rating 5).”}

\end{itemize}

\newpage

\section{Appendix B}
\label{sec:appendix_B}

Total lines where target term appears in the text for both corpora (1970-2016): for the General corpus: mental\_health = 3,233; mental\_illness = 1,559, perception = 9,440; for the Psychology corpus (1970-2016): mental\_health = 26,482; mental\_illness = 4,219, perception = 54,694. 

\begin{figure}[ht]
 \centering
 \includegraphics[width=\columnwidth]{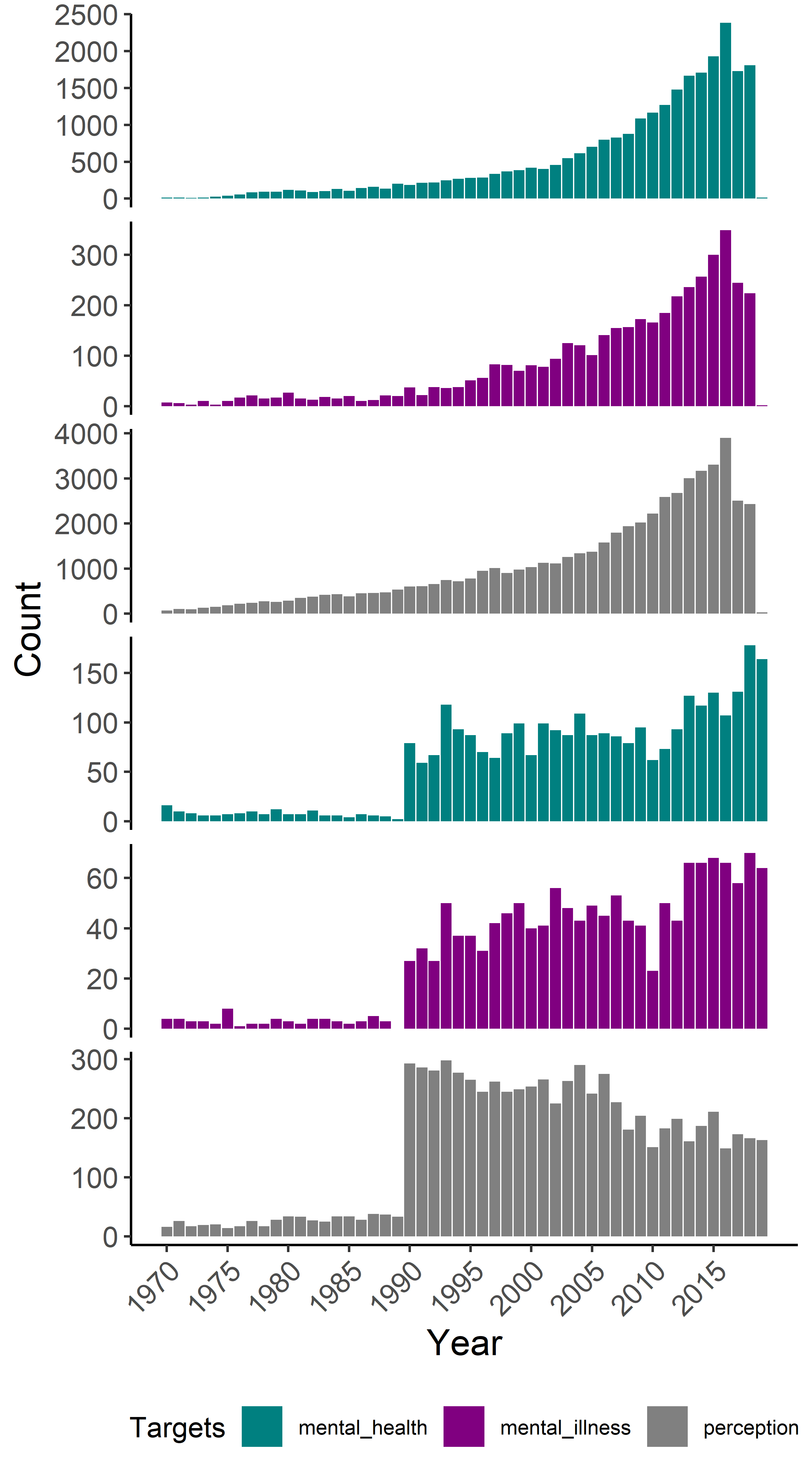}
    \caption{Annual counts of articles where target terms appear in the main text (1970-2016). \emph{Note:} Top three panels = Psychology corpus; bottom three panels = General corpus.}
    \label{fig:AB1}
\end{figure}

\newpage

\section{Appendix C}
\label{sec:appendix_C}

\subsection*{Breadth Model Selection}\
\label{sec:model_select}
The top three (pre-trained) sentence transformer models were chosen, ranked by their performance in embedding sentences.\footnote{\url{https://www.sbert.net/docs/pretrained_models.html}} The best-performing model on the semantic textual similarity benchmark,\footnote{\url{https://paperswithcode.com/sota/semantic-textual-similarity-on-sts-benchmark}} Multi-Task Deep Neural Network \cite{liu2019multi}, was unavailable.\footnote{See \url{https://github.com/namisan/mt-dnn}} See Table~\ref{tab:ACT1} for descriptive statistics of models.

\begin{itemize}
    \item "\textbf{all-mpnet-base-v2}"\footnote{"all-mpnet-base-v2" from Hugging Face, sentence-transformers: \url{https://huggingface.co/sentence-transformers/all-mpnet-base-v2}} is maintained by the SentenceTransformers community and excels in encoding sentences across 14 diverse tasks from different domains using the MPNet (Masked and Permuted Pre-training for Language Understanding) \cite{song2020mpnet} architecture.

\item "\textbf{all-distilroberta-v1}"\footnote{"all-distilroberta-v1" from Hugging Face, sentence-transformers: \url{https://huggingface.co/sentence-transformers/all-distilroberta-v1}} uses a distilled version of "distilroberta-base" \cite{Sanh2019DistilBERTAD}, based on BERT architecture, employing knowledge distillation during pre-training and a triple loss (language modeling, distillation and cosine-distance losses) to leverage the inductive biases of LLMs during pre-training.

\item "\textbf{all-MiniLM-L6-v2}"\footnote{"all-MiniLM-L6-v2": \url{https://huggingface.co/sentence-transformers/all-MiniLM-L6-v2}} uses the MiniLM architecture \cite{wang2020minilm} employing deep self-attention distillation (using self-attention relation distillation for task-agnostic compression of pre-trained Transformers).

\item Additionally, "\textbf{bert-base-uncased}"\footnote{"bert-base-uncased": \url{https://huggingface.co/google-bert/bert-base-uncased}} \cite{Devlin:2019} was included for comparison, although its network structure prohibits the direct comparison of sentence embeddings, and BERT maps sentences to a vector space that is unsuitable for use with common similarity measures and performs below average GloVe embeddings on STS tasks \cite{Reimers&Gurevych:2019}. 
\end{itemize}

\begin{onecolumn}

\begin{table}[!ht]
\centering
\renewcommand{\arraystretch}{1.2} 
\begin{tabular}{p{20mm}p{30mm}p{30mm}p{30mm}p{30mm}} 
 \toprule
\textbf{Model Info} & \textbf{all-mpnet-base-v2*} & \textbf{all-distilroberta-v1*} & \textbf{all-MiniLM-L6-v2*} & \textbf{bert-base-uncased} \\
 \midrule
Accuracy+ & 69.57 & 68.73 & 68.06 & NA \\
Size & 420 MB & 290 MB & 80 MB & 80 MB \\
Case Sensitive & Yes & Yes & Yes & Yes \\
Vocabulary & 30,527 & 50,264 & 30,522 & 30,522 \\
Max Seq Length & 384 & 512 & 256 & 512 \\
\multirow{2}{=}{Pooling} & \multicolumn{3}{p{90mm}}{Mean Pooling (Tokens)} & CLS pooling \\
Dimensions & 768 & 768 & 384 & 768 \\
Layers & 12 & 6 & 6 & 12 \\
Heads & 12 & 12 & 12 & 12 \\
Parameters & 33M & 82.1M & 33M & 110M \\
\multirow{2}{=}{Training Data} & \multicolumn{3}{p{90mm}}{>1B training pairs, sent. (3 data sets: wikihow, code\_search \_net, ms\_marco)} & \multirow{2}{*}{NA} \\
\multirow{2}{=}{Fine-tuning} & \multicolumn{3}{p{90mm}}{\textit{Contrastive Learning Objective}: given a sentence from the sentence pair, the model is trained to predict which out of a set of randomly sampled other sentences, is paired with it in the dataset. It computes the cosine similarity from each possible sentence pair and applies the cross-entropy loss by comparing with true pairs.} & \multirow{2}{*}{NA} \\
 & \multicolumn{3}{p{90mm}}{} & \\
\midrule
 Base Model & mpnet-base & distilroberta-base & MiniLM-L12-H384-uncased & bert-base-uncased \\
Pre-training Corpora & BooksCorpus, CC-News, English Wikipedia, OpenWebText, Stories 
& BooksCorpus, CC-News, English Wikipedia, OpenWebText, Stories
& Unknown (Corpora for the original model used for distillation, UniLMv2, is also unknown)
& Unknown (Likely a large code and text dataset) \\
Pre-training Technique & (1) Permuted language modeling; (2) Incorporate auxiliary positional information 
& (1) Knowledge Distillation, building on the robust training techniques of RoBERTa (dynamic masking, large batch sizes, longer training duration)
& (1) Distillation (deep self-attention distillation) likely from UniLMv2
& (1) Masked language modeling; (2) Next sentence prediction; (3) Tokenization with WordPiece; (4) Positional embeddings \\
 \bottomrule
\end{tabular}
\caption{Summary of language models sampled in the present study. \emph{Note:} * = embeddings are normalized. + = Average performance on encoding sentence over 14 tasks over 14 diverse tasks from different domains (14 datasets). SNL = 570k sentence pairs annotated with labels. Multi-Genre NLI = 430k sentence pairs covering spoken and written text. BookCorpus = 11,038 unpublished books scraped from the Internet.}
\label{tab:ACT1}
\end{table}

\end{onecolumn}

\begin{twocolumn}

\subsection*{Model Comparison: Test Sample}
\label{sec:model_compar}

First, we compared similarity scores for sentence embedding pairs for each sentence transformer model to get a qualitative understanding of the captured dimensions. After feeding seven sample sentences through each sentence transformer model for encoding, similarity arrays of each sentence embedding pair were compared. Tokenization and preprocessing is handled as part of the sentence transformers library. 

\begin{itemize}
    \item 0 = "\textit{She has been seen at a mental\_health facility since 1983.}"
    \item 1 = "\textit{I didn't want to believe I had any mental\_health issues and went into denial.}"
    \item 2 = "\textit{The burden of mental\_illness concentrates in 5-10 of the adolescent population.}"
    \item 3 = "\textit{Their rates of mental\_illness are almost twice that of religious adolescents raised in religious households.}"
    \item 4 = "\textit{Stigma against people with mental\_illness is a very complex public health problem.}"
    \item 5 = "\textit{Stigma associated with mental\_illness is one of the major impediments in evolving effective treatment interventions to address the burden associated with these disorders.}"
    \item 6 = "\textit{Anorexia is a killer it has the highest mortality rate of any mental\_illness, including depression .}"
\end{itemize}

\begin{figure}[ht]
 \centering
 \includegraphics[width=\columnwidth]{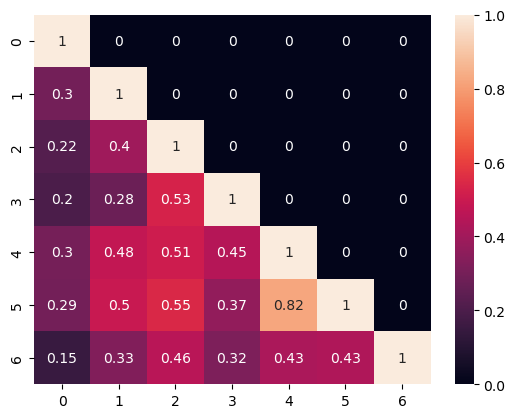}
    \caption{Cosine similarity matrix for sentence embeddings using the "all-mpnet-base-v2" model.}
    \label{fig:AC1}
\end{figure}

 \begin{figure}[ht]
 \centering
 \includegraphics[width=\columnwidth]{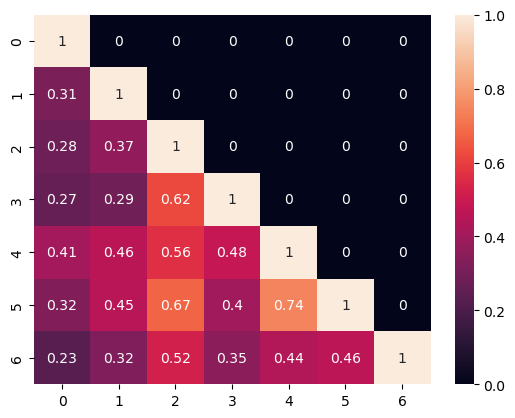}
    \caption{Cosine similarity matrix for sentence embeddings using the "all-distilroberta-v1" model.}
    \label{fig:AC2}
\end{figure}

\begin{figure}[ht]
 \centering
 \includegraphics[width=\columnwidth]{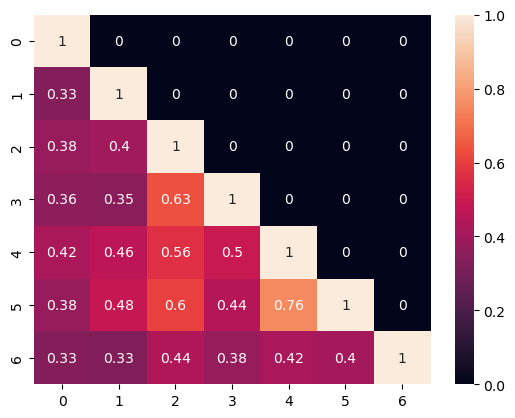}
    \caption{Cosine similarity matrix for sentence embeddings using the "all-MiniLM-L6-v2" model.}
    \label{fig:AC3}
\end{figure}

Our analysis demonstrated that "all-mpnet-base-v2" (the best model on various encoding tasks, as shown in the first row of Table~\ref{tab:ACT1}) had the highest similarity for semantically equivalent sentences (see Figure~\ref{fig:AC1}). 
For this task, its superior performance might be attributed to its architecture. MPNet leverages token dependencies through permuted language modeling, which involves scrambling sentential word order and training the model to predict the original order, forcing MPNet to learn the relationships and dependencies between words. It also incorporates auxiliary positional information, allowing the model to perceive entire sentences, enhancing its ability to capture semantic nuances. "all-distilroberta-v1" (Figure~\ref{fig:AC2}) and "all-MiniLM-L6-v2" (Figure~\ref{fig:AC3}) do not capture this semantic depth, underscoring the strengths of MPNet in semantic understanding and syntactic sensitivity.

\subsection*{Breadth Measure}
\label{sec:breadth_measure}
To analyze semantic differences among sentences containing target concepts, we first extracted texts and then sentences containing target terms from our corpora. The frequency of these sentences over five-year intervals dictated the minimum acceptable number of sentences to sample. 

Next, we engaged in a randomized sampling process. In the 1975-1979 interval, there were more than 50 texts in total, apart from for “mental\_health”, in the general corpus. From these texts, we randomly sampled up to 50 sentences per interval across 10 sets of samples (we sampled all available sentences when there were fewer texts), resulting in up to 500 sentences for every five-year interval, shown in Figure \ref{fig:AC4}.

Following data acquisition, we encoded sentence embeddings using state-of-the-art approaches. Using sentence transformers (except for ``bert-base-uncased'' which tokenized and passed sentences through PyTorch tensors), we derived embeddings that encapsulated the semantic essence of each sentence. These embeddings were averaged along dimension one in the last hidden state layer, creating a single vector representation for each sentence that achieves a nuanced representation of the sentence's semantic content.

Finally, dissimilarity scores were computed. Leveraging the inverse cosine distance metric, we estimated the similarity between every pair of sentence representations using pairwise distances within the range [-1,1]. To ensure unbiased results, we excluded self-similarity and symmetric elements, focusing solely on the upper half of the matrix (49x25). During analysis, the matrix was flattened to extract a 1D array (a stacked half-matrix) of line-by-line similarity scores. Next, we inverted the similarity scores by subtracting them from 1 to obtain absolute values within the range of [0,1], signifying the dissimilarity between corresponding sentence vectors. The final dissimilarity metric was computed by averaging scores within each of the ten samples per interval (getting the sum of cosine distance scores divided by the total number of sentence pairs), followed by an additional averaging across each five-year period within the 1970-2014 range. Higher scores on the cosine distance metric, ranging from 0 to 1, correspond to greater dissimilarity between sentence vectors. 

\begin{figure}[ht!]
 \centering
 \includegraphics[width=\columnwidth]{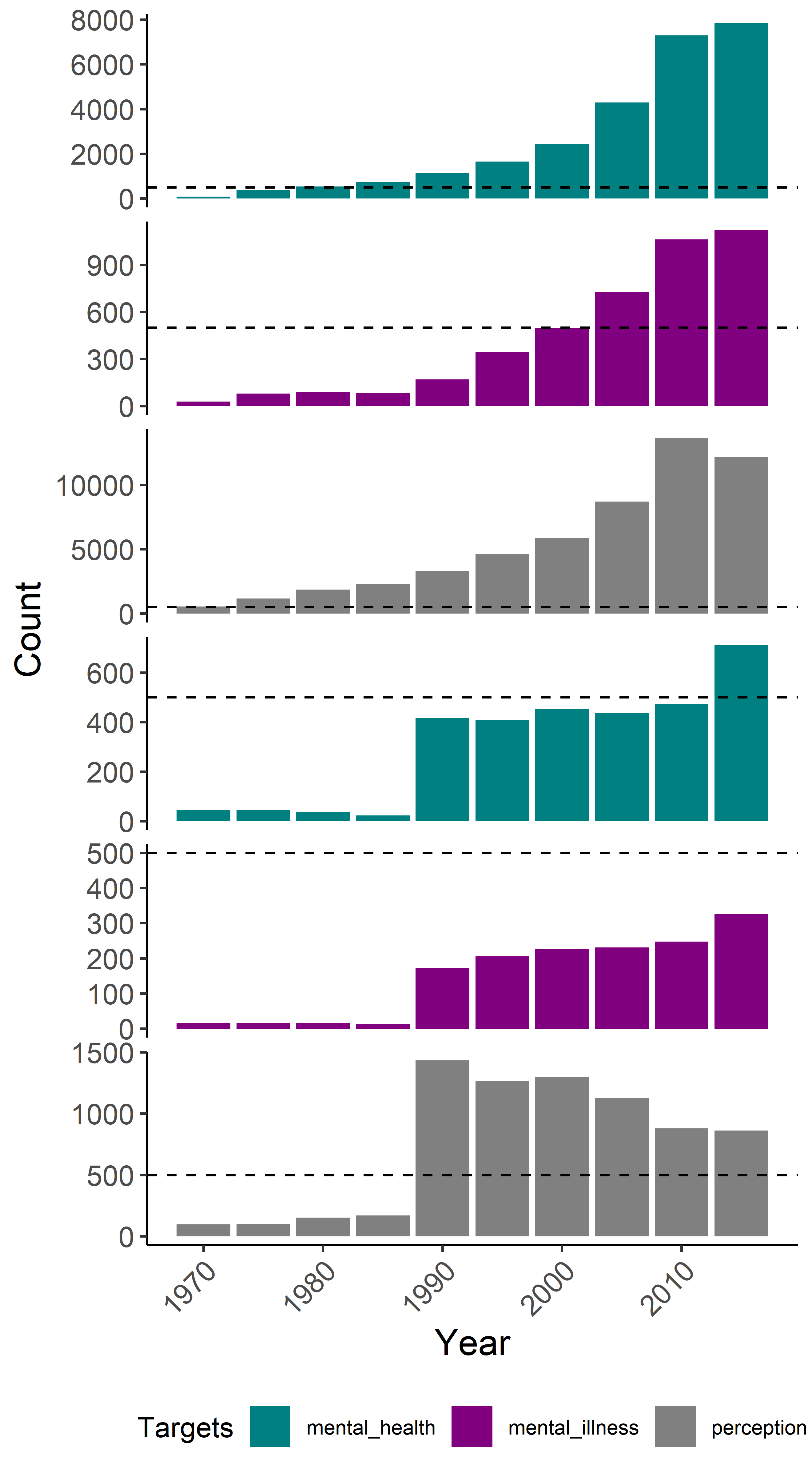}
    \caption{Counts of lines containing target terms grouped by 5-year intervals (horizontal line represents maximum sampling threshold). \emph{Note:} Top three panels = Psychology corpus; bottom three = General corpus.}
    \label{fig:AC4}
\end{figure}

\subsection*{Model Comparison: Results}
\label{sec:model_results}

After computing breadth scores across five-year intervals, we compared trends using sentence transformer models and bert-base-uncased. As shown in Figure~\ref{fig:AC5}, most models showed an upward trend in cosine distance (i.e., inverse similarity), indicating a broader semantic usage of the target concepts. However, "bert-base-uncased" showed lower and flatter similarity scores, possibly due to its pre-training on tasks less directly related to semantic textual similarity. "all-mpnet-base-v2," chosen for the main analysis, performed similarly to the other models but excelled in capturing semantic nuances, as described in Section \ref{sec:model_compar}.

\begin{onecolumn}

\begin{figure*}[!ht]
 \centering
 \includegraphics[width=\textwidth]{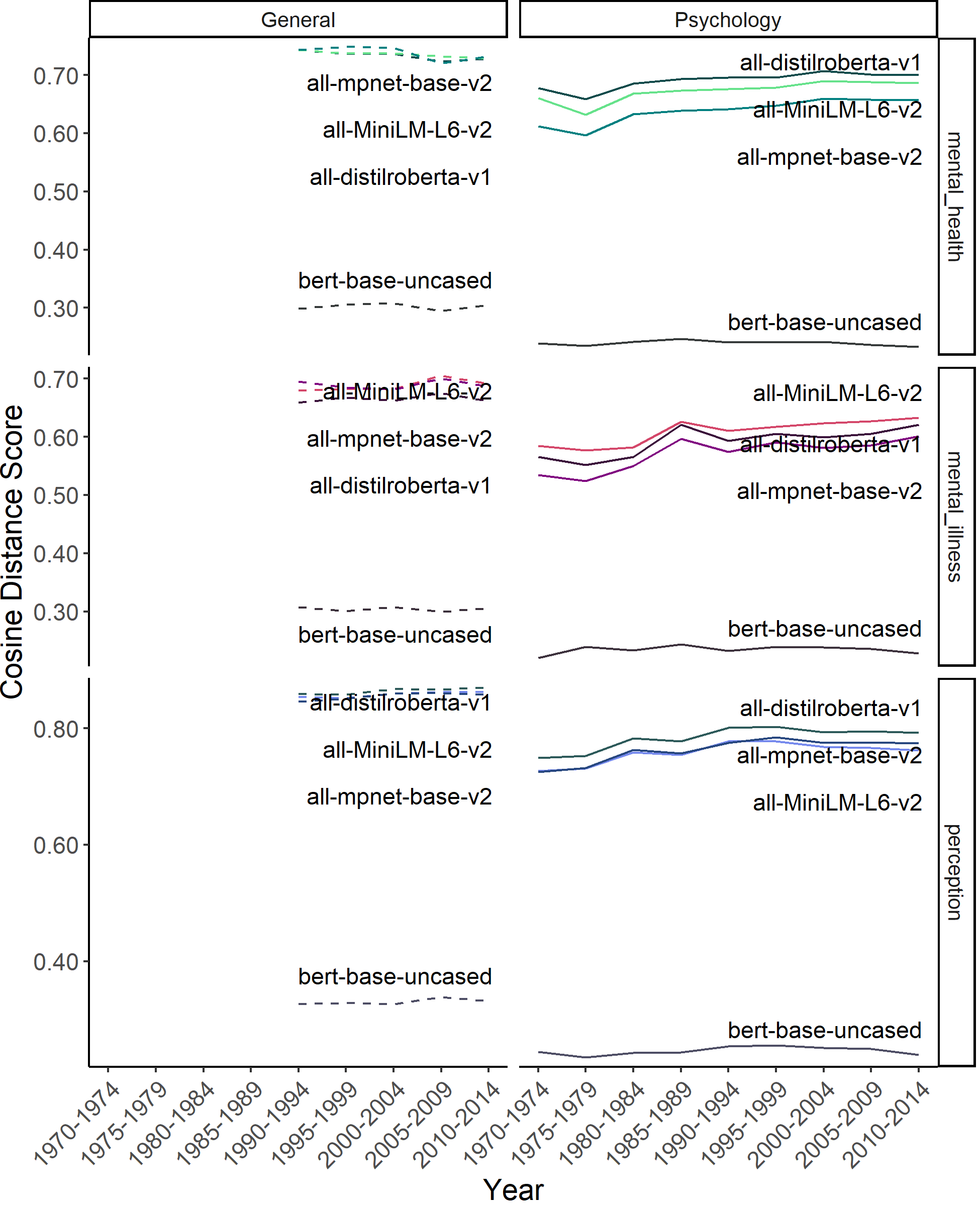}
    \caption{Breadth score over five-year intervals for each model (1970-
2014). \emph{Note:} Model order demonstrates rank of cosine distance score at the final data point (2010-2014) from highest to lowest.}
    \label{fig:AC5}
\end{figure*}

\end{onecolumn}

\begin{twocolumn}

\section{Appendix D}
\label{sec:appendix_D}

To create the general corpus, a rigorous procedure was followed. We first combined two related corpora: the Corpus of Historical American English (CoHA; Davies, 2008) and the Corpus of Contemporary American English (CoCA; Davies, 2008). CoHA contains ~400 million words from 1810-2009, drawn from 115,000 texts distributed across everyday publications (fiction, magazines, newspapers, and non-fiction books). CoCA contains 560 million words from 1990-2019 drawn from ~500,000 texts (from spoken language, TV shows, academic journals, fiction, magazines, newspapers, and blogs). After merging the two corpora, the combined corpus spanning 1810-2019 was processed following recommendations from Alatrash et al. (2020) to clean it without compromising the qualitative and distributional properties of the data. This process included first excluding the special token “@”, which appears in 5\% of the CoHA corpus (introduced for legal reasons), malformed tokens that are possible artifacts of the digitization process or the data processing, and clean-up performed using the web interface (“\&c?;”, “q!”, “|p130”, “NUL”), and removing escaped HTML characters (“ ( STAR ) ”, “<p>”, “<>”). Other symbols were excluded after manual inspection of the corpus (e.g., “ // ”, “ | ”, “ -- ”, “*”, “..”, “PHOTO”, “( COLOR )”, “ ILLUSTRATION ”, “/”). Blogs were also excluded (89,054 web articles; 98,788 blogs) for not containing associated year data, and 25,418 academic texts were removed. Forty-one lines were removed for missing text data (3 fiction, 11 news, 25 magazines, 2 spoken text) and 32 lines were removed for column misalignment (15 mag, 15 news, 1 fiction, 1 tv). The cleaned corpus was then lower-cased and punctuation (commas, periods, question marks), function words, numerals and academic texts were removed. The final combined corpus contained 822,620,111 words from 344,634 texts: 30,496 fiction books, 136,476 magazines, 113,421 newspapers, 2,635 non-fiction books, 43,209 spoken language and 18,397 TV shows. The current study restricted the corpus period from 1970 to 2016 using 501,415,577 tokens from 244,552 articles (23,855 fiction; 88,641 magazines; 73,557 news; 1,498 non-fiction; 40,036 spoken; 16,965 TV). 
\end{twocolumn}

\clearpage

\section{Appendix E}
\label{sec:appendix_E}

\begin{table}[!ht]
\centering
\resizebox{\columnwidth}{!}{
\begin{tabular}{ *{5}{p{16mm}} }
 \toprule
\textbf{1970} & \textbf{1980} & \textbf{1990} & \textbf{2000} & \textbf{2010} \\
 \midrule
positive & poor & poor & poor & poor \\
poor & general & maternal & maternal & positive \\
adolescent & well & positive & well & maternal \\
female & positive & well & positive & well \\
improved & preventive & adolescent & general & adolescent \\
overall & good & general & adolescent & parental \\
preventive & adolescent & good & parental & general \\
public & maternal & bad & bad & bad \\
recent & own & optimal & good & good \\
robust & individual & own & overall & overall \\
 \bottomrule
\end{tabular}
}
\caption{Top 10 adjective modifiers of \emph{mental health} in the psychology corpus (terms are ranked by their relative count for the respective decade)}
\label{tab:table_E1}
\end{table}

\begin{table}[!ht]
\centering
\resizebox{\columnwidth}{!}{
\begin{tabular}{ *{5}{p{14.5mm}} }
 \toprule
\textbf{1970} & \textbf{1980} & \textbf{1990} & \textbf{2000} & \textbf{2010} \\
 \midrule
past & chronic & severe & severe & severe \\
chronic & major & serious & serious & serious \\
excess & severe & chronic & major & chronic \\
feminine & familial & major & chronic & parental \\
formal & malingered & common & parental & major \\
more & acute & maternal & other & common \\
obvious & aggressive & parental & common & other \\
other & disabling & comorbid & maternal & co \\
partum & few & persistent & comorbid & maternal \\
severe & less & other & persistent & comorbid \\
 \bottomrule
\end{tabular}
}
\caption{Top 10 adjective modifiers of \emph{mental illness} in the psychology corpus (terms are ranked by their relative count for the respective decade)}
\label{tab:table_E2}
\end{table}

\begin{table}[!ht]
\centering
\resizebox{\columnwidth}{!}{
\begin{tabular}{ *{5}{p{15.9mm}} }
\toprule
\textbf{1970} & \textbf{1980} & \textbf{1990} & \textbf{2000} & \textbf{2010} \\
\midrule
visual & visual & visual & visual & visual \\
interpersonal & social & social & positive & social \\
auditory & maternal & positive & negative & negative \\
differential & positive & negative & social & positive \\
social & parental & subjective & subjective & conscious \\
subliminal & negative & parental & conscious & subjective \\
pictorial & interpersonal  & maternal & parental & high \\
binocular & human & auditory & high & auditory \\
favorable & subjective  & accurate & categorical  & low \\
high & auditory  & interpersonal  & low & parental \\
\bottomrule
\end{tabular}
}
\label{tab:table_E3}
\caption{Top 10 adjective modifiers of \emph{perception} in the psychology corpus (terms are ranked by their relative count for the respective decade)}
\end{table}

\begin{table}[!ht]
\centering
\resizebox{\columnwidth}{!}{
\begin{tabular}{ *{5}{p{17mm}} }
 \toprule
\textbf{1970} & \textbf{1980} & \textbf{1990} & \textbf{2000} & \textbf{2010} \\
 \midrule
collective & everincreasing & good & good & poor \\
diminished & vibrant & own & well & good \\
necessary & NA & positive & own & well \\
normal & NA & rural & collective & abysmal \\
NA & NA & sound & optimal & additional \\
NA & NA & subsequent & poor & comprehensive \\
NA & NA & bad & postpartum & lessthanoptimal \\
NA & NA & dubious & collegestudent & new \\
NA & NA & geriatric & confident & own \\
NA & NA & maternal & fragile & pediatric \\
 \bottomrule
\end{tabular}
}
\label{tab:table_E4}
\caption{Top 10 adjective modifiers of \emph{mental health} in the general corpus (terms are ranked by their relative count for the respective decade)}
\end{table}

\begin{table}[!ht]
\centering
\resizebox{\columnwidth}{!}{
\begin{tabular}{ *{5}{p{13.5mm}} }
 \toprule
\textbf{1970} & \textbf{1980} & \textbf{1990} & \textbf{2000} & \textbf{2010} \\
 \midrule
serious & acute & severe & severe & serious \\
certain & hereditary & serious & serious & severe \\
incipient & severe & major & major & other \\
socalled & socalled & chronic & other & acute \\
NA & underlying & untreated & most & chronic \\
NA & NA & common & adolescent & common \\
NA & NA & other & bipolar & diagnosable \\
NA & NA & classic & common & major \\
NA & NA & more & many & deep \\
NA & NA & severe & new & difficult \\
 \bottomrule
\end{tabular}
}
\caption{Top 10 adjective modifiers of \emph{mental illness} in the general corpus (terms are ranked by their relative count for the respective decade)}
\label{tab:table_E5}

\end{table}

\begin{table}[!ht]
\centering
\resizebox{\columnwidth}{!}{
\begin{tabular}{ *{5}{p{15.5mm}} }
 \toprule
\textbf{1970} & \textbf{1980} & \textbf{1990} & \textbf{2000} & \textbf{2010} \\
 \midrule
extrasensory & public & public & public & public \\
visual & different & widespread & common & common \\
own & common & visual & popular & popular \\
aesthetic & general & common & wrong & general \\
direct & innate & popular & human & own \\
single & own & general & general & sensory \\
keen & popular & own & own & veridical \\
new & psychic & new & extrasensory & extrasensory \\
practical & clear & extrasensory & acute & negative \\
present & human & human & visual & parental \\
 \bottomrule
\end{tabular}
}
\caption{Top 10 adjective modifiers of \emph{perception} in the general corpus (terms are ranked by their relative count for the respective decade)}
\label{tab:table_E6}
\end{table}

\clearpage

\section{Appendix F}
\label{sec:appendix_F}

\begin{table}[!ht]
\centering
\resizebox{\columnwidth}{!}{
\begin{tabular}{ *{5}{p{17.5mm}} }
 \toprule
\textbf{1970} & \textbf{1980} & \textbf{1990} & \textbf{2000} & \textbf{2010} \\
 \midrule
community & service & service & service & service \\ 
center & community & child & problem & problem \\ 
service & professional & professional & child & child \\ 
program & center & use & use & study \\ 
professional & problem & problem & care & use \\ 
child & use & care & study & care \\ 
school & study & study & treatment & treatment \\ 
problem & social & treatment & professional & need \\ 
group & child & need & need & outcome \\ 
worker & program & community & health & physical \\ 
 \bottomrule
\end{tabular}
}
\label{tab:table_F1}
\caption{Top 10 Warriner-matched collocates of \emph{mental health} in the psychology corpus (terms are ranked by their relative count for the respective decade)}
\end{table}

\begin{table}[!ht]
\centering
\resizebox{\columnwidth}{!}{
\begin{tabular}{ *{5}{l} }
 \toprule
\textbf{1970} & \textbf{1980} & \textbf{1990} & \textbf{2000} & \textbf{2010} \\
 \midrule
attitude & attitude & severe & severe & people \\
scale & study & person & people & severe \\
patient & patient & patient & use & study \\
study & high & treatment & patient & stigma \\
group & person & substance & study & use \\
psychiatric & problem & study & person & individual \\
opinion & scale & use & disorder & treatment \\
factor & child & people & treatment & disorder \\
find & use & disorder & individual & family \\
student & major & family & substance & experience \\
 \bottomrule
\end{tabular}
}

\label{tab:table_F2}
\caption{Top 10 Warriner-matched collocates of \emph{mental illness} in the psychology corpus (terms are ranked by their relative count for the respective decade)}
\end{table}

\begin{table}[!ht]
\centering
\resizebox{\columnwidth}{!}{
\begin{tabular}{ *{5}{p{15mm}} }
 \toprule
\textbf{1970} & \textbf{1980} & \textbf{1990} & \textbf{2000} & \textbf{2010} \\
 \midrule
self & study & study & study & study \\
study & child & self & self & self \\
result & self & child & child & social \\
test & result & social & examine & relationship \\
visual & subject & result & relationship & use \\
child & difference & examine & use & examine \\
subject & social & use & social & influence \\
group & relationship & relationship & result & effect \\
difference & effect & difference & relate & result \\
use & group & behavior & influence & child \\
 \bottomrule
\end{tabular}
}
\label{tab:table_F3}
\caption{Top 10 Warriner-matched collocates of \emph{perception} in the psychology corpus (terms are ranked by their relative count for the respective decade)}
\end{table}

\begin{table}[!ht]
\centering
\resizebox{\columnwidth}{!}{
\begin{tabular}{ *{5}{p{14.8mm}} }
 \toprule
\textbf{1970} & \textbf{1980} & \textbf{1990} & \textbf{2000} & \textbf{2010} \\
 \midrule
department & center & have & say & have \\
state & institute & national & have & say \\
center & service & institute & national & issue \\
health & fund & care & institute & care \\
city & have & service & child & problem \\
director & national & professional & care & system \\
institute & allow & abuse & community & service \\
national & commissioner & state & need & health \\
new & department & center & service & physical \\
program & oak & department & problem & professional \\
 \bottomrule
\end{tabular}
}
\label{tab:table_F4}
\caption{Top 10 Warriner-matched collocates of \emph{mental health} in the general corpus (terms are ranked by their relative count for the respective decade)}
\end{table}

\begin{table}[!ht]
\centering
\resizebox{\columnwidth}{!}{
\begin{tabular}{ *{5}{p{12.5mm}} }
 \toprule
\textbf{1970} & \textbf{1980} & \textbf{1990} & \textbf{2000} & \textbf{2010} \\
 \midrule
drug & suffer & have & have & have \\
history & alcoholism & people & people & people \\
treat & have & severe & family & family \\
acute & time & depression & do & say \\
appoint & acute & family & suffer & alliance \\
bill & argue & say & disorder & history \\
can & ask & can & say & national \\
cancer & basement & drug & severe & severe \\
cause & bout & history & child & suffer \\
center & cite & know & drug & member \\
 \bottomrule
\end{tabular}
}
\label{tab:table_F5}
\caption{Top 10 Warriner-matched collocates of \emph{mental illness} in the general corpus (terms are ranked by their relative count for the respective decade)}
\end{table}

\begin{table}[!ht]
\centering
\resizebox{\columnwidth}{!}{
\begin{tabular}{ *{5}{p{12.3mm}} }
\toprule
\textbf{1970} & \textbf{1980} & \textbf{1990} & \textbf{2000} & \textbf{2010} \\ \midrule
people & public & public & change & change \\
public & change & change & public & public \\
alter & reality & can & people & people \\
change & base & people & can & can \\
president & black & other & reality & time \\
study & member & reality & other & shift \\
associate & new & world & go & affect \\
can & people & may & depth & alter \\
cause & popular & go & know & challenge \\
child & side & thing & will & pain \\
\bottomrule
\end{tabular}
}
\label{tab:table_F6}
\caption{Top 10 Warriner-matched collocates of \emph{perception} in the general corpus (terms are ranked by their relative count for the respective decade)}
\end{table}

\end{twocolumn}

\begin{onecolumn}

\section{Appendix G}
\label{sec:appendix_G}

\begin{figure}[!ht]
 \centering
 \includegraphics[width=\columnwidth]{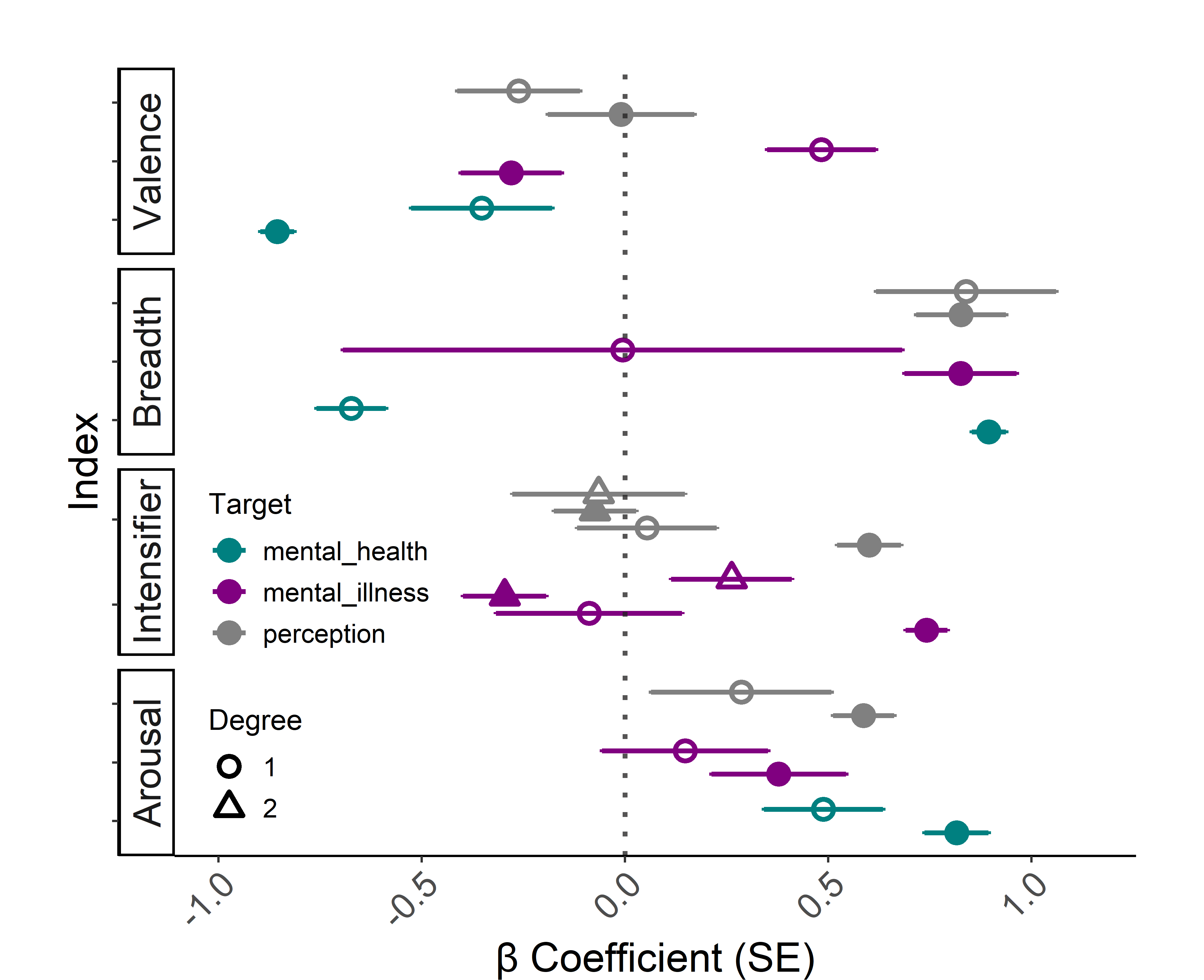}
    \caption*{Year effect sizes for indices operationalizing major dimensions of lexical semantic change in the psychology corpus (filled circles) and general corpus (empty circles). \emph{Note:} First degree = Linear; Second degree = Quadratic. Vertical dotted line = Standardized beta coefficient of 0; Standard errors (SE) that overlap line indicate that the null hypothesis can be rejected at the 5\% significance level.}
    \label{fig:AG1}
\end{figure}

\begin{table*}[!ht]
    \centering
    \begin{tabular}{llllllll}
        \toprule
        \textbf{Index (Concept)} & \textbf{Corpus} & \textbf{Model} & \textit{\textbf{B}} & \textbf{SE} & \textbf{\textit{t}} & \textbf{\textit{p}} & \textbf{\textit{F} (DF)}; \textbf{Adj.\textit{ R\textsuperscript{2}}}  \\
        \midrule
        \multirow{4}{*}{Intensifier (\emph{Mental Illness})} & \multirow{2}{*}{Psychology} & Linear & 0.74 & 0.09 & 8.18 & $<$.001 & 38.76 (1, 44); 0.62* \\
         & & Quadratic & -0.33 & 0.10 & -3.26 & 0.002  \\
        \cmidrule{2-8}
        \multirow{4}{*}{} & \multirow{2}{*}{General} & Linear & -0.09 & 0.20 & -0.45 & .657 & 0.99 (2, 24); -0.0005 \\
        & & Quadratic & 0.30 & 0.22 & 1.34 & 0.194 \\
        \midrule
        \multirow{4}{*}{Intensifier (\emph{Perception})} & \multirow{2}{*}{Psychology} & Linear & 0.60 & 0.12 & 5.00 & $<$.001 & 12.71 (2,44); 0.34* \\
         & & Quadratic & -0.08 & 0.14 & -0.62 & 0.541 \\
        \cmidrule{2-8}
        \multirow{4}{*}{} & \multirow{2}{*}{General} & Linear & 0.05 & 0.20 & 0.27 & 0.793 & 0.08 (2, 24); -0.08 \\
        & & Quadratic & -0.07 & 0.23 & -0.32 & 0.752 \\
        \bottomrule
    \end{tabular}
    \caption{Regression Coefficients (Scaled) and Fit Statistics Predicting Intensifier Indices as a Function of Year. \emph{Note}: * = \emph{p}-value for the overall model = <.001. Regression coefficients are unstandardized. For mental\_illness in psychology, residuals were autocorrelated, and outcome variable was re-fit with Generalized Least Squares approach, yielding: B = 0.74; SE = 0.09; \textit{p} < .001; RSE(DF) = 0.62(47,44); BIC = 108.52.}
    \label{tab:table_AG1}
\end{table*}

\begin{table*}[!ht]
\centering
\begin{tabular}{llllllll}
\toprule
\textbf{Index} & \textbf{Concept} & \textbf{Corpus} & \textit{\textbf{B}}   & \textbf{SE}  & \textbf{\textit{p}} & \textbf{\textit{F} (DF)} & \textbf{Adj. \textit{R\textsuperscript{2}}} \\
\midrule
\multirow{6}{*}{Valence} & \multirow{2}{*}{\emph{Mental Health}} & Psychology & -0.003 & \num{3e-4} & <.001 & 122.65 (1,45) & 0.73 \\ \cmidrule{3-8}
                            &                                        & General    & -0.005 & 0.003  & .071 & 3.55 (1,25) & 0.09    \\
\cmidrule{2-8}
                            & \multirow{2}{*}{\emph{Mental Illness}} & Psychology & -0.002 & \num{9e-4} & .057  & 3.82 (1,45) & 0.058   \\  \cmidrule{3-8}
                            &                             & General    
                            & 0.01  & 0.005  & .011 & 7.62 (1,25) & 0.20 \\
\cmidrule{2-8}
                            & \multirow{2}{*}{\emph{Perception}} & Psychology & \num{-1e-5} & \num{2e-4} & .949  & 0.004 (1,45) & -0.02  \\  \cmidrule{3-8}
                            &                                        & General    & -0.002 & 0.002 & .188 &  1.84 (1,25) & 0.03 \\
\midrule
\multirow{6}{*}{Breadth}    & \multirow{2}{*}{\emph{Mental Health}} & Psychology & 0.001 & \num{3e-4} & 0.001 & 28.19 (1,7) & 0.77   \\  \cmidrule{3-8}
                            &                                       & General    & -0.001 & \num{7e-4}  & .213 & 2.49 (1,3) & 0.27  \\
\cmidrule{2-8}
                            & \multirow{2}{*}{\emph{Mental Illness}} & Psychology & 0.002  & \num{4e-4}  & .006 & 14.99 (1,7) & 0.64    \\  \cmidrule{3-8}
                            &                                        & General    & \num{-6e-6}  & \num{6e-4}   & .992 & \num{1e-4} (1,3) & -0.33     \\
\cmidrule{2-8}
                            & \multirow{2}{*}{\emph{Perception}} & Psychology & 0.001 & \num{3e-4} & 0.006 & 15.12 (1,7) & 0.64  \\  \cmidrule{3-8}
                            &                                        & General    & \num{7e-4}   & \num{3e-4}  & .076 & 7.13 (1,3) & 0.61 \\
\midrule
\multirow{6}{*}{Arousal}    & \multirow{2}{*}{\emph{Mental Health}} & Psychology & 0.003 & \num{3e-4} & <.001 & 89.38 (1,45) & 0.66   \\  \cmidrule{3-8}
                            &                                       & General    & 0.005 & 0.002 & <.001 & 7.83 (1,25) & 0.21  \\
\cmidrule{2-8}
                            & \multirow{2}{*}{\emph{Mental Illness}} & Psychology & 0.003  & \num{9e-4}  & <.001 & 7.51 (1,45) & 0.12    \\  \cmidrule{3-8}
                            &                                        & General    & 0.002  &  0.003 & .462 & 0.56 (1,25) & -0.02     \\
\cmidrule{2-8}
                            & \multirow{2}{*}{\emph{Perception}} & Psychology & 0.001 & \num{2e-4} & <.001 & 23.65 (1,45) & 0.33  \\  \cmidrule{3-8}
                            &                                        & General    & 0.002   & 0.001 & .148 & 2.22 (1,25) & 0.05 \\
\midrule
\midrule
\multirow{6}{*}{Path.} & \multirow{2}{*}{\emph{Mental Health}} & Psychology & \num{4e-4} & \num{3e-5} & <.001 & 163.34 (1,45) & 0.78 \\  \cmidrule{3-8}
                                  &                                        & General    & \num{3e-4} & \num{2e-4} & .130 & 2.48 (1,21) & 0.06  \\
\cmidrule{2-8}
                                  & \multirow{2}{*}{\emph{Mental Illness}} & Psychology & \num{2e-4} & \num{1e-4} & .049 & 4.12 (1,43) & 0.07 \\  \cmidrule{3-8}
                                  &                                        & General    & \num{-1e-4} & \num{2e-4}  & .552  & 0.36 (1,23) & -0.03  \\
\cmidrule{2-8}
                            & \multirow{2}{*}{\emph{Perception}} & Psychology & \num{2e-3} & \num{4e-2} & <.001 & 118.42 (1,44)  & 0.72 \\  \cmidrule{3-8}
                            &                                        & General    & \num{5e-5}  & \num{2e-5}  & .051 & 5.95 (1,6) & 0.41 \\
\midrule
\multirow{6}{*}{Salience}    & \multirow{2}{*}{\emph{Mental Health}} & Psychology & \num{7e-6} & \num{4e-7} & <.001 & 292.52 (1,45) & 0.86\\  \cmidrule{3-8}
                              &                                        & General    & \num{2e-7} & \num{4e-8} & <.001 & 18.17 (1,45) & 0.27 \\
\cmidrule{2-8}
                              & \multirow{2}{*}{\emph{Mental Illness}} & Psychology & \num{3e-7} & \num{9e-8} & <.001 & 13.21 (1,45) & 0.21 \\  \cmidrule{3-8}
                              &                                        & General    & \num{1e-7} & \num{2e-8} & <.001 & 42.21 (1,45) & 0.47 \\
\cmidrule{2-8}
                            & \multirow{2}{*}{\emph{Perception}} & Psychology & \num{5e-7} & \num{3e-7} & .160 & 2.04 (1,45) & 0.02 \\  \cmidrule{3-8}
                            &                                    & General     & \num{-3e-8}  & \num{6e-8}  & .568  & 0.33 (1,45) & -0.01 \\
\bottomrule
\end{tabular}
\caption{Unstandardized Regression Coefficients and Fit Statistics Predicting Indices as a Function of Year. {\emph{Note}: The midrule separates the main dimensions (above) and the exploratory dimensions (below). Path. = Pathologization. Generalized Least Squares approach also used for models with autocorrelated residuals. \\
• Arousal: \textit{mental\_health} (P): B = 0.003; SE = \num{3e-4}; \textit{p} < .001; RSE(DF) = 0.03(47,45); BIC = -172.07 \\
• Salience: \textit{mental\_health} (P): B = \num{7e-6}; SE = \num{4e-7}; \textit{p} <.001; RSE(DF) = \num{4e-5}(47,45); BIC = -767.87; \textit{mental\_illness} (P): B = \num{3e-7}; SE = \num{9e-7}; \textit{p} < .001; RSE(DF) = \num{9e-6}(47,45); BIC = -895.27; \textit{perception} (P): B = \num{5e-7}; SE = \num{3e-7}; \textit{p} = .160; RSE(DF) = \num{3e-5} (47,45); BIC = -785.60; \textit{mental\_illness} (G): B = \num{1e-7}; SE = \num{2e-8}; \textit{p} < .001; RSE(DF) = \num{2e-6}(47,45); BIC = -1048.85
\label{tab:table_AG2}
}}
\end{table*}

\newpage

\vspace{-5cm}
\begin{table*}[htbp]
\centering
 \begin{tabular}{p{0.15\linewidth}lllll}
   \toprule
   \textbf{Index} & \textbf{Concept} & \textbf{Corpus} & \textbf{$\beta$} & SE & \textbf{95\% CI} \\
   \midrule
   \multirow{6}{*}{Valence} & \multirow{2}{*}{\emph{Mental Health}} & Psychology & -0.86\textsuperscript{*} & 0.04 & (-0.94, -0.77) \\  \cmidrule{3-6}
                             &                                       & General & -0.35 & 0.17 & (-0.71, 0.004) \\
   \cmidrule{2-6}
                             & \multirow{2}{*}{\emph{Mental Illness}} & Psychology & -0.28\textsuperscript{*} & 0.12 & (-0.53, -0.03) \\  \cmidrule{3-6}
                             &                                       & General & 0.48\textsuperscript{*} & 0.13 & (0.21, 0.76) \\
   \cmidrule{2-6}
                             & \multirow{2}{*}{\emph{Perception}} & Psychology & -0.01 & 0.18 & (-0.37, 0.35) \\  \cmidrule{3-6}
                             &                                       & General & -0.26 & 0.15 & (-0.57, 0.05) \\
   \midrule
   \multirow{6}{*}{Breadth} & \multirow{2}{*}{\emph{Mental Health}} & Psychology & 0.90\textsuperscript{*} & 0.04 & (0.80, 0.99) \\  \cmidrule{3-6}
                             &                                       & General & -0.67\textsuperscript{*} & 0.09 & (-0.95, -0.40) \\
   \cmidrule{2-6}
                             & \multirow{2}{*}{\emph{Mental Illness}} & Psychology & 0.83\textsuperscript{*} & 0.14 & (0.50, 1.15) \\  \cmidrule{3-6}
                             &                                       & General & -0.01 & 0.69 & (-2.19, 2.18) \\
   \cmidrule{2-6}
                             & \multirow{2}{*}{\emph{Perception}} & Psychology & 0.83\textsuperscript{*} & 0.11 & (0.57, 1.09) \\  \cmidrule{3-6}
                             &                                       & General & 0.84\textsuperscript{*} & 0.22 & (0.13, 1.54) \\
   \midrule
   \multirow{8}{*}{Intensifier} & \multirow{4}{*}{\emph{Mental illness}} 
   
             & Psychology(1) & 0.74* & 0.05 & (0.64, 0.85) \\
          &  & Psychology(2) & -0.30* & 0.10 & (-0.50, -0.09) \\  \cmidrule{3-6}
          &  & General(1) & -0.09 & 0.23 & (-0.56, 0.38) \\
          &  & General(2) & 0.26 & 0.15 & (-0.05, 0.57) \\
\cmidrule(lr){2-6}
          & \multirow{4}{*}{\emph{Perception}} 
          
             & Psychology(1) & 0.60* & 0.08 & (0.44, 0.76) \\
          &  & Psychology(2) & -0.07 & 0.10 & (-0.28, 0.13) \\  \cmidrule{3-6}
          &  & General(1) & 0.05 & 0.17 & (-0.30, 0.41) \\
          &  & General(2) & -0.06 & 0.21 & (-0.50, 0.37) \\
   \midrule
   \multirow{6}{*}{Arousal} & \multirow{2}{*}{\emph{Mental Health}} & Psychology & 0.82* & 0.08 & (0.66, 0.97) \\  \cmidrule{3-6}
                             &                                       & General & 0.49 & 0.15 & (0.19, 0.79) \\  
   \cmidrule{2-6}
                             & \multirow{2}{*}{\emph{Mental Illness}} & Psychology & 0.38\textsuperscript{*} & 0.17 & (0.05, 0.71) \\  \cmidrule{3-6}
                             &                                       & General & 0.15 & 0.20 & (-0.27, 0.57) \\  
   \cmidrule{2-6}
                             & \multirow{2}{*}{\emph{Perception}} & Psychology & 0.59* & 0.08 & (0.44, 0.74) \\  \cmidrule{3-6}
                             &                                       & General & 0.29 & 0.22 & (-0.17, 0.74) \\
   \midrule
   \midrule
   \multirow{6}{*}{Pathologization} & \multirow{2}{*}{\emph{Mental Health}} & Psychology & 0.30* & 0.12 & (0.06, 0.53) \\  \cmidrule{3-6}
                             &                                       & General & -0.12 & 0.23 & (-0.61, 0.36) \\ 
   \cmidrule{2-6}
                             & \multirow{2}{*}{\emph{Mental Illness}} & Psychology & 0.89* & 0.02 & (0.85, 0.92) \\  \cmidrule{3-6}
                             &                                       & General & 0.32 & 0.20 & (-0.09, 0.74) \\
   \cmidrule{2-6}
                             & \multirow{2}{*}{\emph{Perception}} & Psychology & 0.85* & 0.30 & (0.79, 0.92) \\  \cmidrule{3-6}
                             &                                       & General & 0.71 & 0.30 & (-0.03, 1.45) \\
   \midrule
   \multirow{6}{*}{Salience} & \multirow{2}{*}{\emph{Mental Health}} & Psychology & 0.93* & 0.02 & (0.89, 0.97) \\    \cmidrule{3-6}
                             &                                       & General & 0.54* & 0.10 & (0.34, 0.73) \\
   \cmidrule{2-6}
                             & \multirow{2}{*}{\emph{Mental Illness}} & Psychology & 0.48* & 0.13 & (0.21, 0.74) \\  \cmidrule{3-6}
                             &                                       & General & 0.70* & 0.07 & (0.56, 0.83) \\
   \cmidrule{2-6}
                             & \multirow{2}{*}{\emph{Perception}} & Psychology & 0.21 & 0.15 & (-0.10, 0.52) \\  \cmidrule{3-6}
                             &                                       & General & -0.09 & 0.15 & (-0.38, 0.21) \\
   \midrule
 \end{tabular}
   \caption{Standardized Regression Coefficients (\textbf{$\beta$}) predicting Semantic Change Indices by Year. \textit{Note}: Midrule separates main dimensions of semantic change (above). * = \emph{p}: < .05. (1) = First degree. (2) = Second degree.} 
   \label{tab:table_AG3}
\end{table*}

\end{onecolumn}

\end{document}